\documentclass[10pt,journal,compsoc]{IEEEtran}
\usepackage{amsmath}
\usepackage{amsfonts}
\usepackage{amssymb}
\usepackage{amsthm}
\usepackage{booktabs}
\usepackage{epstopdf}
\usepackage{enumerate}
\usepackage{hyperref}
\usepackage[switch]{lineno}
\usepackage{multirow}
\usepackage{pdfpages}
\usepackage{rotating}
\usepackage{subfigure}
\usepackage{tabularx}
\usepackage{textcomp}
\usepackage{xcolor}
\usepackage{graphicx}
\usepackage[algo2e,linesnumbered,lined,boxed,commentsnumbered,ruled]{algorithm2e} 
\usepackage{url}
\usepackage{ragged2e}
\usepackage[justification=centering]{caption}
\usepackage{cite}
\usepackage{caption}
\newtheorem{definition}{Definition}

\begin{document}


\title{Open Continual Feature Selection via \\ Granular-Ball Knowledge Transfer}

\author{Xuemei Cao, 
        Xin Yang*, ~\IEEEmembership{Member,~IEEE},
        Shuyin Xia, ~\IEEEmembership{Member,~IEEE},
        Guoyin Wang, ~\IEEEmembership{Senior Member,~IEEE},
        Tianrui Li, ~\IEEEmembership{Senior Member,~IEEE}
\thanks{Xuemei Cao and Xin Yang are with the School of Computing and Artificial Intelligence, Southwestern University of Finance and Economics, Chengdu 611130, China. E-mail: caoxuemei.qpz@gmail.com, yangxin@swufe.edu.cn. \\
Shuyin Xia and Guoyin Wang are with the Chongqing Key Laboratory of Computational Intelligence, Chongqing University of Posts and Telecommunications, Chongqing 400065, China. E-mail: xiasy@cqupt.edu.cn, wanggy@cqupt.edu.cn. \\
Tianrui Li is with School of Computing and Artificial Intelligence, Southwest Jiaotong University, Chengdu 611756, China. E-mail:trli@swjtu.edu.cn. \\
Corresponding authors: Xin Yang.

}
}

\IEEEtitleabstractindextext{

\justifying

\begin{abstract}

This paper presents a novel framework for continual feature selection (CFS) in data preprocessing, particularly in the context of an open and dynamic environment where unknown classes may emerge. CFS encounters two primary challenges: the discovery of unknown knowledge and the transfer of known knowledge. To this end, the proposed CFS method combines the strengths of continual learning (CL) with granular-ball computing (GBC), which focuses on constructing a granular-ball knowledge base to detect unknown classes and facilitate the transfer of previously learned knowledge for further feature selection. CFS consists of two stages: initial learning and open learning. The former aims to establish an initial knowledge base through multi-granularity representation using granular-balls. The latter utilizes prior granular-ball knowledge to identify unknowns, updates the knowledge base for granular-ball knowledge transfer, reinforces old knowledge, and integrates new knowledge. Subsequently, we devise an optimal feature subset mechanism that incorporates minimal new features into the existing optimal subset, often yielding superior results during each period. Extensive experimental results on public benchmark datasets demonstrate our method's superiority in terms of both effectiveness and efficiency compared to state-of-the-art feature selection methods.



\end{abstract}

}

\IEEEdisplaynontitleabstractindextext
\IEEEpeerreviewmaketitle

\maketitle

\section{Introduction}

With the rapid advancement of data acquisition and storage technologies, the datasets generated in practical applications are often large-scale and high-dimensional.
Ubiquitously, these datasets encompass redundant or irrelevant features, which not only exacerbate the challenges inherent in learning task modeling, but also may lead to poor learning performance and model interpretability.
Therefore, how to reduce the dimensionality of such datasets to improve the performance of subsequent tasks has become an urgent problem to be solved immediately in various practical applications.

Feature selection, or attribute reduction, is an effective technique to mitigate the above problem.
Its essence lies in finding an `optimal feature subspace' by removing unimportant features from the original feature space without degrading the performance of the learning model and at the same time achieving the purpose of dimensionality reduction.
At present, feature selection has attracted significant interest in academia and has been widely utilized in clustering analysis, classification learning, approximate reasoning, and data mining \cite{zhu2020dgdfs}.
Over the past three decades, a variety of feature selection methods have been developed, and their research focus has changed as technology has advanced \cite{you2023robust,maji2012rough}. 
Earlier studies primarily sought to identify a more precise feature subspace for specific tasks.
In recent years, there has been a growing interest in the dynamic nature of real-world data, with attempts to address the challenges posed by the continuous growth of samples, the continuous addition of features, and the constant change in feature values.
However, the success of all these methods depends on the assumption that all class labels are known and presented to the learner before feature selection takes place.
This implies that neither of these methods encountered any unseen classes during either the training or testing phases. 

This assumption provides a simplified abstraction that enables complex tasks to be addressed in a more tractable manner, leading to the flourishing development of feature selection techniques.
It bears emphasizing that the real world represents an open environment subject to constant change.
Compared to the static data of the closed environment in the past, the open environment will not only increases known instances but also instances of unknown classes.
Accordingly, the assumptions of traditional methods are invalid in many practical scenarios where not all class labels can be present in advance.
For example, consider the scenario of forest disease monitoring aided by a machine learning model whose performance is enhanced by feature selection on signals sent from numerous sensors deployed in the forest \cite{zhou2022open}.
Since environmental changes can cause new forest diseases to emerge, such as those caused by a new invasive pest, it is infeasible for humans to enumerate all possible disease class labels at the beginning of modeling.

An interesting question is, if we use high-level computers to generate these classes in advance, should we develop new methods to make reasonable assumptions about the feature values of the new classes and compute them, or should we wait a long time until all the classes are known and then implement the learning process?
Unfortunately, even if users have knowledge about it, it is impossible to predict which classes will emerge in the future and when they will manifest.
At the same time, retaining all past data for an indefinite duration becomes prohibitively costly and infeasible when new classes emerge at a substantially delayed point in time.
Obviously, these solutions will lead to high storage costs and expensive computation times and are thus not suitable for practical applications.
Furthermore, when open data arrives in the form of streams, existing feature selection methods must learn all the data from scratch and fail to leverage previously learned knowledge, incurring redundant computational costs and slowing response to new data.
It is worth noting that for privacy protection reasons, historical data may not be accessible once hidden, making it impossible to combine old and new data to obtain comprehensive training data.

Ideally, a more pragmatic way would be to transfer previously learned knowledge to new periods, guiding rapid learning in new data.
The focus should be on improving or updating the trained feature selection model based solely on new data, avoiding starting from scratch.
Regrettably, model updates with new data can lead to the well-known issue of losing previously acquired knowledge, known as catastrophic forgetting.
To mitigate this, a new learning paradigm, Continual Learning (CL), has been recently proposed \cite{aljundi2019task,mundt2023wholistic}.
CL aims to facilitate knowledge transfer and prevent knowledge forgetting across a sequence of tasks.
However, most current CL studies focus on predictive tasks in neural networks and lack methodologies for feature selection.
Compared to the representation of knowledge via network parameters in CL, the depiction of knowledge in feature selection is observed to be more challenging. 
To address this complexity, the introduction and development of innovative theories and mechanisms are required.

Granular-ball computing (GBC) is a new sample space representation method that is efficient, robust, and has good knowledge representation capabilities.
It has garnered increasing attention from researchers and has been successfully applied in classification \cite{xia2022accurate}, clustering \cite{xie2023efficient}, and feature selection \cite{xia2020gbnrs}.
Although a recent innovation, GBC's concept is rooted in the large-scale priority characteristics and multi-granularity cognitive mechanisms of humans, first discovered in 1982.
The primary objective of GBC is the construction of an optimal set of granular balls, focusing on coverage, purity, and quantity.
Yet, its application in scenarios involving unknown class addition in open environments has not been studied, making it a worthwhile topic for investigation.

Based on the aforementioned observations, CL is introduced into feature selection to avoid redundant computations through knowledge transfer.
A novel framework for continual feature selection based on GBC as the foundational model is proposed.
This framework comprises four modules: the base model, class identification, granular-ball updating, and feature subset enhancement.
Each module addresses a specific learning challenge: constructing the initial knowledge base, identifying open sets, transforming unknowns into knowns, and rapidly updating feature subsets, thereby enabling continual feature selection in open environments.
The main contribution of this paper can be summarized as follows:

(1) To the best of our knowledge, this is the first attempt to formulate and study a feature selection problem in an open environment where known and unknown classes dynamically emerge.
More critically, for practical challenges such as the inability to repeatedly access historical data, the knowledge base proposed in this paper, which synthesizes granular-balls and selected feature subsets, demonstrates greater intelligibility and universality.

(2) A novel framework for continual feature selection on open data streams is proposed.
It consists of four modules designed for the initial learning of knowledge discovery and the open learning of knowledge transfer.
It effectively recognizes and learns both known and unknown classes, with its foundation on granular balls enhancing efficiency and robustness.

(3) Comprehensive experiments are performed on multiple publicly available datasets covering two particle sphere models. Compared with existing methods, our method has better generality and effectiveness, and can perform continuous feature selection well in the open world.

The rest of this paper is organized as follows. 
In Section 2, some related feature selection methods and related notions are briefly reviewed. 
Section 3 presents the basic model and its definitions.
The details of the proposed method are presented in Section 4.
Section 5 verifies the effectiveness and efficiency of the proposed method. 
Section 6 concludes the paper.

\section{Related Work}

Feature selection is an effective data preprocessing technique in machine learning that aims to reduce data dimensionality by deleting redundant features, thereby improving model performance.
The study of feature selection has a long history and is generally considered to be a problem of searching for an optimal subset of features \cite{wang2017feature,roffo2020infinite}.
According to the form of the training data, feature selection can be broadly classified into two categories: static and dynamic.
Static methods focus on close-environments and assume that important factors in the learning process hold invariant, including the features describing training and testing data never change, the number of data instances used for training and testing never dynamically increase, and all data drawn from the same distribution \cite{Wang2023Outliers}.
Based on this assumption, research on static feature selection methods has flourished, and many such methods have been proposed in the last three decades.
However, data for real-world applications is full of dynamic changes, so there is an increasing need for feature selection methods that go beyond the limitations of closed-environments.

\begin{figure}[!htb]
\centering
\includegraphics[width=\linewidth]{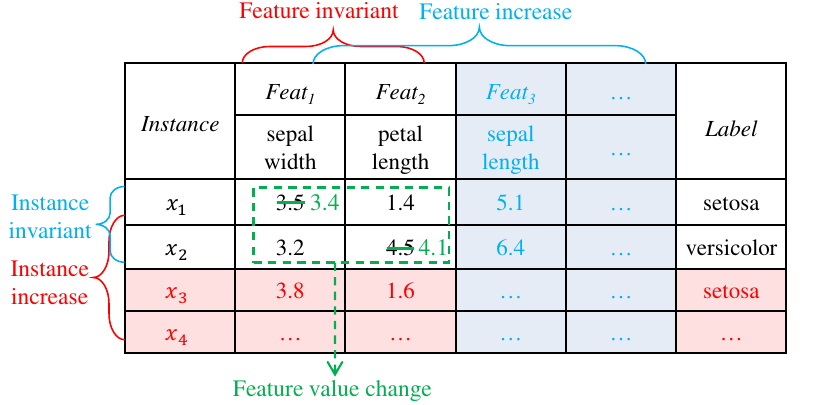}
\caption{There are three dynamic changes in the data, and the colors red, blue, and green represent one of them.}
\label{fig: 3fs}
\end{figure}

Fortunately, with the development of technology, more and more dynamic feature selection has been proposed recently to deal with the changes of important factors in the learning process.
As illustrated in Fig. \ref{fig: 3fs}, dynamic methods can be divided into three distinct groups based on changes in three aspects: instances, features, and feature values.
The first group is for the variation of instances, which assumes that the features describing the data are fixed while the number of instances is continuously increasing.
During the preliminary phase of the research, the instance varied in a single increasing manner \cite{liu1999incremental, chen2016incremental}.
For example, Liu et al. \cite{liu1999incremental} proposed an algorithm for finding the minimal reduct, which is suitable for information systems without decision labels. 
To address the time complexity issues arising from incremental data, Chen et al. \cite{chen2016incremental} presented a feature selection method grounded in variable precision rough set theory.
Further research faces the dynamic increase of instance sets with multiple instances \cite{liang2012group, zhang2019active, yang2017incremental}.
For example, Liang et al. \cite{liang2012group} developed an algorithm that uses information entropy to measure feature significance and selects significant features as the final feature subset. 
Zhang et al. \cite{zhang2019active} investigated the information entropy incremental mechanisms based on fuzzy rough set and then presented an active incremental feature selection algorithm with incoming instances.
Yang et al. \cite{yang2017incremental} addressed the issue of data increment by dynamically adding and deleting features based on relative discernibility relations and developed two incremental algorithms for fuzzy rough set feature selection.

The second group is for the variation of features, which makes the exact opposite assumptions of the first group.
It focuses on scenarios where the number of training instances is fixed, while the volume of features grows over time. 
For example,  Li et al. \cite{li2007rough} designed a dynamic method for updating approximations of rough sets based on the characteristic relation in an incomplete information system.
Based on three representative entropies, Wang et al. \cite{wang2013attribute} proposed a dynamic feature selection algorithm under the mechanism of increasing the dimensionality of data features.
From the perspective of incrementally updating knowledge granulation, Qian et al. \cite{qian2016feature} presented an incremental feature selection algorithm when a feature set is simultaneously added to and deleted from the system.
In addition, several researchers have explored online stream feature selection when features arrive as streams over time \cite{liu2023asfs, you2021online, yan2022online}.
For instance, Liu et al. \cite{liu2023asfs} proposed an online streaming feature selection method for multi-label learning based on multiple objectives.
You et al. \cite{you2021online} developed an online streaming feature selection method that considers label correlation for multi-label scenarios.

The third group is for the variation of feature values, which assumes that the number of instances and features are invariant while the feature values change dynamically \cite{wang2013reduction, shu2014incremental, xie2018novel, chen2014rough}.
For example, Wang et al. \cite{wang2013reduction} investigated the properties of entropy when adding features to information systems, proposed an entropy update mechanism, and then developed a feature selection algorithm for feature value changes.
After updating the positive domain by an incremental method, Shu and Shen \cite{shu2014incremental} presented two efficient incremental feature selection algorithms to deal with the different dynamic changes in feature values of single instance and multiple instances.
By introducing the concept of inconsistency degree in incomplete decision systems, Xie et al. \cite{xie2018novel} investigated an update mechanism with three strategies and further proposed an incremental feature selection algorithm framework for dynamic incomplete decision systems.
Chen et al. \cite{chen2014rough} introduced the minimal discernibility attribute set to address attribute value variations due to coarsening and refining in reduction and approximation processes. They also developed a rough set-based decision rule update method for inconsistent decision systems, enhancing feature selection efficiency.

The works discussed above share a common assumption that all class labels are available from the beginning and do not change dynamically over time.
Unfortunately, this does not always hold.
As with the forest disease monitor case mentioned previously, real-world applications exist in open environments, where not only known class instances but also unknown class instances will appear continuously.
Consequently, this poses great challenges to existing feature selection methods, and how to build a reasonable framework for continuous feature selection to accommodate open environments is still an unsolved problem.
To this end, we propose a continual feature selection framework for an open world where unknown classes emerge.


\section{Base Model and Definition}  
As the base model in the CFS framework, the granular ball neighborhood rough set is very important. Here, we will introduce its related concepts and mathematical underpinnings in detail.

\subsection{Granular-Ball Computing}
\begin{figure}[ht]
    \centering
    \subfigure[First iteration.]{
        \includegraphics[width=0.20\textwidth]{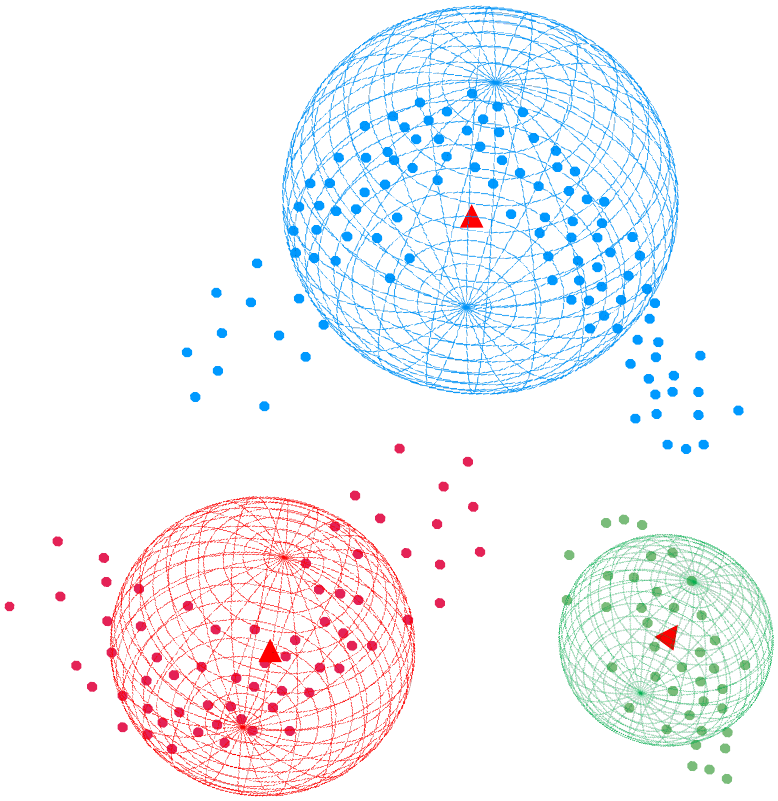}
        \label{fig:GB_iteration1}
     }
    \hspace{8pt}
    \subfigure[Last iteration.]{
        \includegraphics[width=0.22\textwidth]{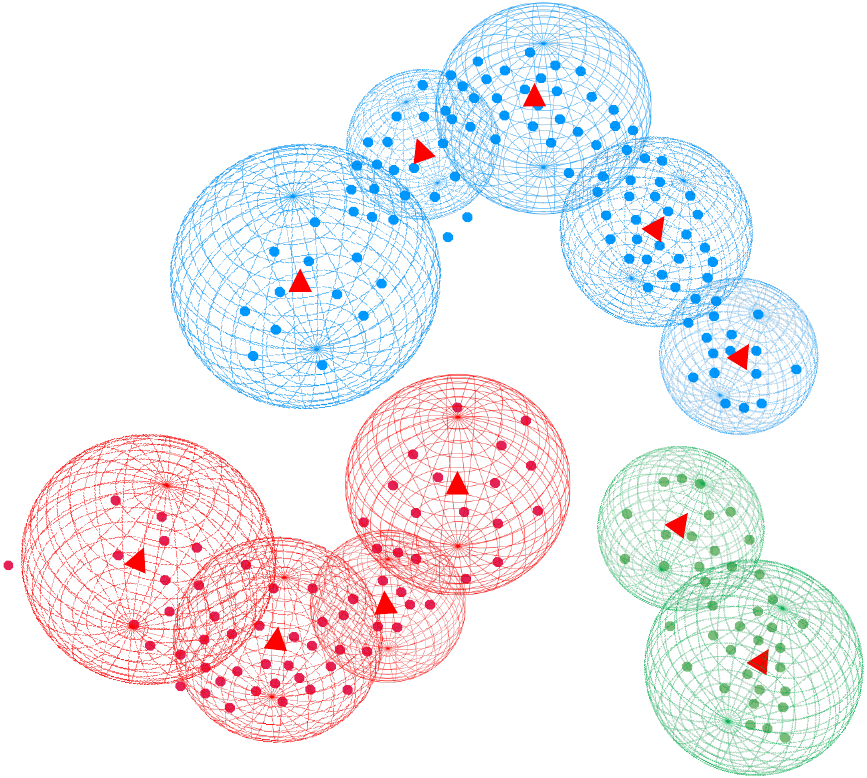}
        \label{fig:GB_iteration2}
    }
    \caption{Granular-balls generation process}
    \label{fig:GB_iteration}
\end{figure}

Granular-ball computing (GBC) was proposed to address the issue that most current learners use a single instance as the finest granularity input, which compromises the efficiency, robustness, and interpretability of their learning process \cite{xia2022efficient}. 
The core concept of GBC lies in the generation and representation of granular-balls, which adaptively employs granular-balls of varying sizes to represent and cover the sample space.
Therefore, GBC can fit arbitrarily distributed data \cite{xie2023efficient}.
Reference \cite{xia2023granular} summarizes various methods for generating granular-balls, with the most common approach being based on splitting according to K-means. 
An intuitive process for generating a type of granular-ball is depicted in Fig. \ref{fig:GB_iteration}, where the dataset consists of three classes, and the initial clustering centers are the centers of each class.
Fig. \ref{fig:GB_iteration1} and Fig. \ref{fig:GB_iteration2} represent the results of the first and last iterations, respectively.
The following is a formal definition of GBC.

Given a dataset $U = \{x_i, i = 1, 2, \dots, n\}$, where $x_i$ and $n$ represent the instance and the number of instances in $U$, respectively.
$G = \{GB_1, GB_2, \dots, GB_m\}$ is a set of granular-balls generated on $U$, and the $j^{th}$ granular-ball $GB_j =\{x_i | i = 1, 2, \dots, k\}$.
The standard model for GBC is shown below:
\begin{equation}\label{granular-ball}
\begin{gathered}
f(x,\vec{\alpha})\longrightarrow g(GB,\vec{\beta}) \\
s.t.\quad \begin{array}{cc}\min&\frac{n}{\sum_{i=1}^{k}|GB_{i}|}+m+loss(GB)\end{array}  \\
s.t. \quad quality(GB_i)\geq T,
\end{gathered}
\end{equation}
where $T$ is the threshold.
$f(x, \vec{\alpha})$ and $g(GB, \vec{\beta})$ represent the existing learning models that take point $x$ and granule $GB$ as input respectively, where $\vec{\alpha}$ and $\vec{\beta}$ are model parameters.
$\sum_{j=1}^m |GB_j|$ represents the coverage degree of data $U$, $m$ represents the total number of granular-balls, and $loss(GB)$ is the information loss.
By constraining these three key factors, the optimal granular-ball generation results can be obtained to minimize the value of the whole equation.

Each granular-ball has only two parameters, center and radius, making it easy to characterize data in any dimension.
Specifically, for each $GB_j$, the center $c_j$ and radius $r_j$ are defined as follows:
\begin{equation}\label{center}
    c_j = \sum_{i=1}^{n_j} \xi_i p_j,
\end{equation}
\begin{equation}\label{radius}
    r_j = \max (distance(p_i,c_j)),
\end{equation}
where $n_j$ represents the number of data points located in $GB_j$, and $\xi$ denotes the weight coefficient.
It is worth noting that the specific weight coefficients and distance functions may be different in different studies.
In \cite{xia2022accurate}, $\xi = \frac{1}{n_j}$, and Euclidean distance is adopted as the distance function.
In \cite{fang2022hypersphere}, $\xi$ is equal to the Lagrange multipliers, and the Gaussian kernel is used as the distance function.

\subsection{Granular-Ball Neighborhood Rough Sets and Feature Selection}

Granular-ball neighborhood rough set was proposed to solve the problem of low efficiency of the traditional neighborhood rough set grid search for the optimal neighborhood radius \cite{xia2020gbnrs}. It was also developed by mapping the low-dimensional feature space to the high-dimensional space, introducing a novel approach that streamlines granular generation and enhances granular stability \cite{fang2022hypersphere}.

Let the triplet $\langle U, A, D \rangle$ be the decision table, where $U=\{x_1, x_2, \dots, x_n\}$ denotes a non-empty finite set of objects, and $U$ is called the universe, $A=\{a_1, a_2, \dots, a_m\}$ denotes a non-empty finite set of features, and $D$ denotes a non-empty finite set of class labels.
The related concepts of granular-ball neighborhood rough set are as follows:

Let $\langle U, A, D \rangle$ be the decision table, $B \subseteq A$, and let $G = \{GB_1, GB_2, \dots, GB_m\}$ be the set of spheres generated by covering $U$, $c_j$ and $r_j$ are the center and radius of the $GB_j$ respectively.
For $x_i \in {GB_j}$, the neighborhood of $x_i$ on $B$ is as follows:
\begin{equation}\label{neighborhood}
    R_B(x_i) = \{x|\forall x \in GB_j, distance_B(x, c_j)\leq {r_j}\},
\end{equation}
where $distance_B(x, c_j)$ is the distance from $x$ to $c_j$ under the feature set $B$, which is consistent with the distance function in Eq. \eqref{radius}.

Let $\langle U, A, D \rangle$ be the decision table, $U/D = \{E_1, E_2, \dots, E_L\}$, $B \subseteq A$, and $GB_j$ be the j-th granular ball.
The generation lower approximation set of $X$ with respect to a feature subset $B$ is defined as:
\begin{equation}\label{gl_approximation_X}
    \underline{B} E' = \{x =\sum_{k=1}^{l_j} \xi_k x_k | x_k \in GB_j(B), R_B(x_k) \subseteq E_i\}.
\end{equation}
According to the Eq. \eqref{gl_approximation_X}, the generation lower approximation of the decision feature set $D$ with respect to a feature set $B$ is defined as:
\begin{equation}\label{gl_approximation_D}
    \underline{B}D' = \bigcup_{i=1}^L \underline{B}E_i',
\end{equation}
where $\underline{B}E_i'=\{x =\sum_{k=1}^{l_j} \xi_k x_k | x_k \in GB_j(B), R_B(x_k) \subseteq E_i\}$.
Therefore, the generation positive region is defined as:
\begin{equation}\label{generation_positive_region}
    GPos_B(D) = \underline{B}D'.
\end{equation}
If the generation positive domain remains unchanged after deleting a feature, then this feature can be considered as a relative redundant feature. The specific mathematical description is as follows:
\begin{equation}\label{relative_redundant_feature}
    GPos_{B-\{a\}}(D) = GPos_B(D).
\end{equation}

Interestingly, the Eq. \eqref{gl_approximation_D} can be understood intuitively as the generation lower approximation consists of the center of a granular-ball containing only one class label.
As described in Eq. \eqref{center}, the centers can be virtual.
In other words, the objects in the generation positive domain are generated rather than selected from existing instances.
According to Eq. \eqref{relative_redundant_feature}, feature selection can be achieved by sequentially examining the entire feature set $A$.
To ensure the accuracy of feature selection, granular-balls must be rebuilt each time a relative redundant feature is deleted.



\section{The Proposed Approach}
In this section, we first elaborate on the research motivation.
Then, we formulate the problem of continual feature selection in an open world.
After that, we present a detailed description of the proposed approach, including the overall framework and the four key modules.

\subsection{Motivation}
Current feature selection methods assume full knowledge of class labels, a condition rarely met in practical settings.
When confronted with data containing unknown classes, existing methods not only fail, but the feature subsets they select may not be optimal.
For instance, in a dataset distinguishing between `Dog' and `Bird' (as illustrated in Table \ref{table: example_motivation}), certain features, like `Jump,' may be redundant, emphasizing the need for feature selection to enhance classifier performance.
Thus, selecting pivotal features (e.g., `Mammal') becomes crucial for enhancing classifier efficiency, as it represents the most relevant feature subset.
We assume that Mammal is the only feature selected by the evaluation function, and $\{Mammal\}$ is the current optimal feature subset.

\begin{figure}[!htb]
\centering
\includegraphics[width=\linewidth]{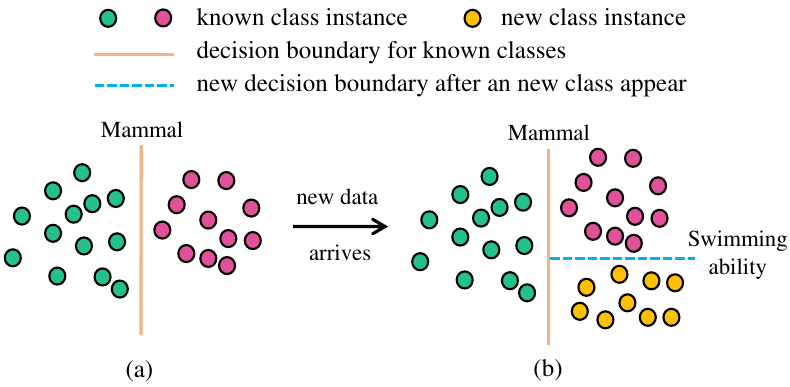}
\caption{Motivation of CFS. After adding new class data, a new feature subset is needed to represent the decision boundary.}
\label{fig: motivation}
\end{figure}

Actually, each feature can be viewed as a partition in the data space.
The current decision boundary, illustrated in Fig. \ref{fig: motivation}(a), uses the 'Mammal' feature to distinguish between the `Dog' and `Bird' classes, achieving accurate classification.
However, the introduction of new classes, such as `Swan' in Fig. \ref{fig: motivation}(b), alters the data distribution and challenges this boundary.
While `Mammal' can differentiate `Dog' from `Swan', it fails to separate `Bird' and `Swan'. 
To address this, we incorporate `Swimming Ability' as an additional feature, effectively distinguishing between `Bird' and `Swan'. 
Consequently, the optimal feature subset for the classifier is now ${\text{Mammal, Swimming Ability}}$.

\begin{table}[!htb]
\renewcommand\arraystretch{1.5}
\caption{Decision Table for Animal}
\label{table: example_motivation}
\centering \arraybackslash
\resizebox{0.96\linewidth}{!}{
\begin{tabular}{| c|c|c|c|c|c|cccccc}
\hline
Instance  &Wing  &Jump &Mammal &Swimming Ability  & Class \\
\hline                                      
$x_1$    &Yes    & Yes & Yes   &  Yes             & Dog   \\
\hline
$x_2$    &No     & Yes & No    &  No              & Bird  \\
\hline
\end{tabular}}
\end{table}

Drawing from the aforementioned observations, we can reveal two phenomena:
(1) The arrival of new classes often reduces the classification accuracy of a classifier using a previously optimal feature subset. 
(2) Enhancing this subset with additional features can improve classification accuracy for both new and existing classes.
Addressing these issues in a dynamic setting highlights the importance of continuous feature selection methods, a concept that has not been considered in previous studies.
In view of this, we propose a continuous feature selection framework based on granular-ball knowledge transfer in an open environment.
It introduces a continual learning paradigm and builds a knowledge base, thereby enhancing the model's adaptability and robustness in handling streaming data.

\subsection{Problem Formulation}
To clearly describe the problem, we provide a formal definition of continual feature selection problem in the open world.
We consider two kinds of data with $d$ feature dimensions: one is the labeled initial data $U_0$, whose class label is a set of $l$ known elements $D = \{y_1, y_2, \dots, y_l\}$.
The other is the unlabeled data stream over $T$ consecutive time periods, $U=\langle U_1, U_2, \dots, U_T\rangle$.
$S_0$ and $G_0$ are the optimal feature subset and granular-ball set obtained on the $U_0$ respectively.
In the time period $t = 1, 2, \dots, T$, we need to identify the unlabeled data $D_t$ arriving for the current time period, and then rapidly calculate the optimal feature subset selection $S_t$ for the current all data $D_t' = \bigcup_{i=0}^{t} D_i$ on the basis of $S_{t-1}$ and $G_{t-1}$.

For each instance $x \in U_t$, our approach aims to identify whether it is a known or unknown class.
If it is a known class, it will be assigned the label $y(x) \in D$.
Otherwise, $x$ is learned further to get a pseudo-label $y'(x) \notin D$.
Furthermore, for each class $y \in D \cup \{y^{pseudo-label}\}$, in order to constantly obtain the optimal feature subset in the future, its granular-ball set $G_{t-1}$ is required to be updated to $G_{t}$.

\subsection{Framework}

\begin{figure*}[!htb]
\centering
\includegraphics[width=0.97\linewidth]{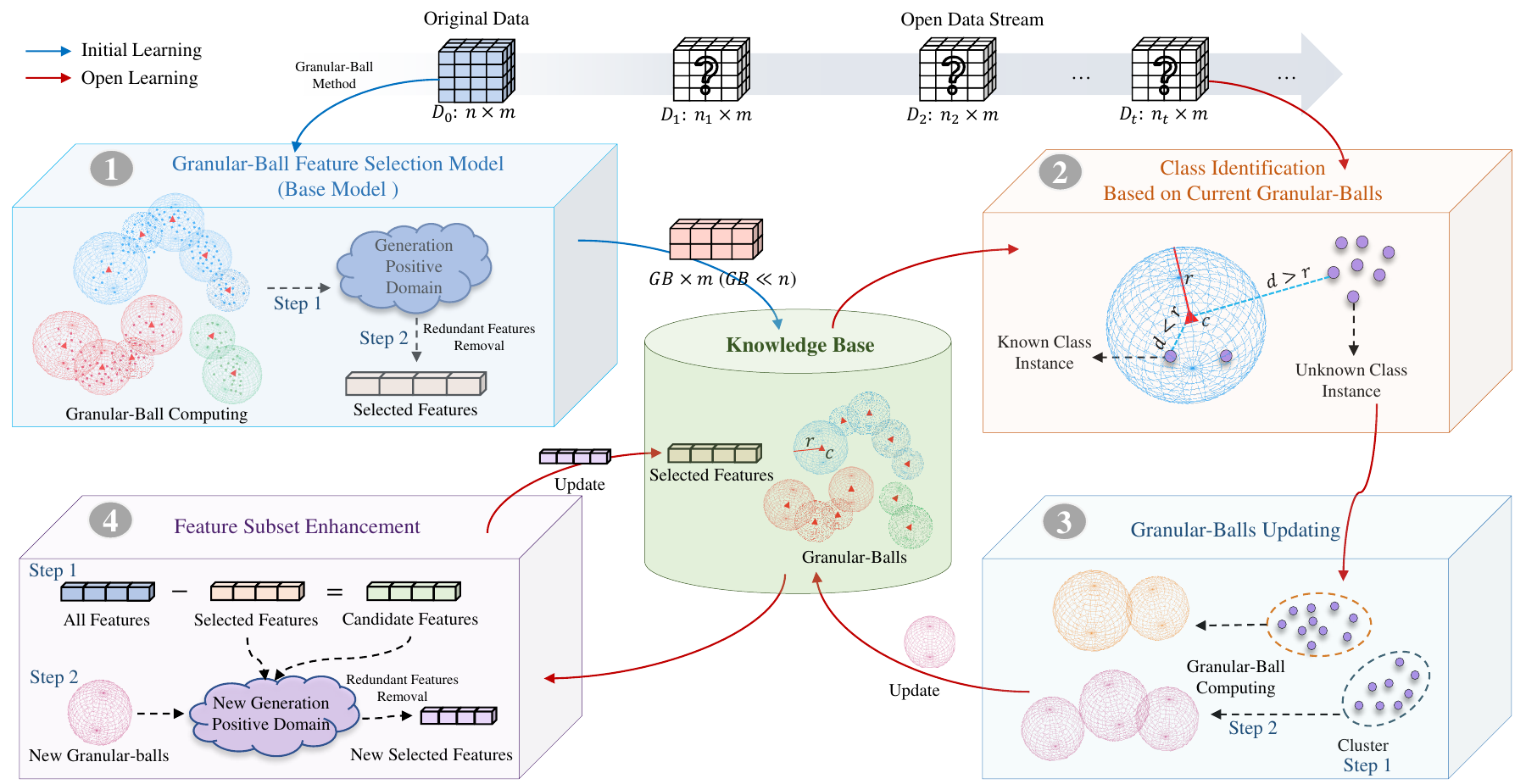} 
\caption{Framework of the Proposed Approach CFS.}
\label{fig: framework}
\end{figure*}

With the granular-ball basic model of feature selection, we propose a continual feature selection framework, as shown in Fig. \ref{fig: framework}.
It comprises four key modules: Base Model, Class Identification, Granular-ball Updating, and Feature Subset Enhancement.
In the initial stage, the Base Model module generates a set of granular-balls and feature subsets.
To facilitate effective knowledge transfer in open-environment learning, we establish a knowledge base using these initial elements. 
This knowledge base acts as supervisory information within the open environment to control risks in the open space.
\begin{definition}\label{knowlwdge_base}(Initial Knowledge Base)
    Given an original dataset $U_0$, $G = \{GB_1, GB_2, \dots, GB_m\}$ is a set of granular-balls generated on $U_0$. Let $S_0$ be the optimal feature subset found on $U_0$. 
    The initial knowledge base $KB$ on $U_0$ is defined as follows:
    \begin{equation}\label{knowlwdge_base}
    KB = \{G, S_0\}.
    \end{equation}
\end{definition}

Upon receiving new data, our framework first activates the Class Identification module.
This step involves analyzing the new data's position relative to the granular-balls in the knowledge base to classify it as known or unknown.
Different learning strategies are then applied: known classes undergo knowledge consolidation, while unknown classes trigger the Granular-ball Updating module.
Here, clustering transforms unknown classes into known ones. 
Subsequently, granular-ball computing generates and updates new granular-balls based on clustered data, ensuring the knowledge base remains current. 
Finally, the Feature Subset Enhancement module recalculates the optimal feature subset for this new period, updating the knowledge base to enhance future learning responsiveness and performance.

\subsection{Class Identification}

In dynamic, open environments, the continuous emergence of unlabeled data, comprising both known and unknown classes, makes traditional supervised, static models unsuitable for feature selection due to a lack of label information. 
Unknown class instances, in particular, offer knowledge beyond the model's current scope, significantly improving accuracy and generalization.
Therefore, promptly and accurately identifying data in these streams is crucial for quick adaptation to environmental changes and guiding model updates.
An ideal model should accurately classify known class instances while also detecting and reporting new, unseen instances among unknown class data.

Technically, class identification can be seen as a process similar to anomaly detection.
The challenge here is to distinguish between anomalies relative to known classes and new instances of unknown classes.
Practically, it is often valid to consider unknown class instances as more ``anomalous" than those from known classes, as suggested in \cite{mu2017classification}. 
Additionally, the volume of genuine anomalies is typically much smaller than that of normal data.
Another significant challenge is leveraging existing knowledge for class identification efficiently without the need for extra anomaly detection modules that would complicate the basic model.


As mentioned before, the granular-balls constructed on the initial data $U_0$ can clearly describe the distribution of the current data.
Their coarse-grained nature, less influenced by fine-grained instance points, contributes to their robustness. 
Consequently, these granular-balls are utilized as prior knowledge to assist in class identification.
Building on the established concepts of granular-balls, we provide definitions for known and unknown class instances in new data:

\begin{definition}\label{known_and_unknown}(Known Class and Unknown Class)
    Given a dataset $U_t=\{x_1, x_2, \dots, x_n\}$ of time period $t$ and an original dataset $U_0$, $G = \{GB_1, GB_2, \dots, GB_m\}$ is a set of granular-balls generated on $U_0$. 
    For each $GB_j$, $c_j$ and $r_j$ are the center and radius of $GB_j$ respectively.
    The \textit{Known Class} and \textit{Unknown Class} in $U_t$ are defined as follows:
    \begin{equation}\label{known_class}
     KC = \{x | \forall x \in U_t, dis(x, c_j) \leq r_j, j = 1, 2, \dots, m\},
    \end{equation}
    \begin{equation}\label{unknown_class}
     UC = \{x | \forall x \in U_t, dis(x, c_j) > r_j, j = 1, 2, \dots, m\},
    \end{equation}
where $dis()$ is the same distance function as $distance()$ in Eq. \eqref{radius}, and $m$ is the number of all granular-balls.
\end{definition}
As outlined in Definition \ref{known_and_unknown}, we determine the class of each instance in $U_t$ by assessing its position relative to the granular-balls. 
This involves calculating the distance of instance $x$ from the center of each granular-ball and comparing these distances to the respective granular-ball radii.
If the distance to all granular-balls exceeds their radii, $x$ is classified as an unknown class instance.
Conversely, if $x$ is within the radius of any granular ball, it is identified as a known class instance, adopting the class label of that granular-ball.

\subsection{Granular-Ball Updating}
Upon identifying known class instances $KC$ and unknown class instances $UC$, it is crucial to update the model to enhance its adaptability for future data.
The challenge lies in enhancing adaptability without compromising the performance of known classes.
Employing experts to label new data for merging and retraining would incur significant labeling, computational, and storage costs.
A more efficient approach involves localized refinements to accommodate new data, avoiding extensive global alterations.
To this end, we have devised distinct processing strategies for known and unknown class instances, aiming to both reinforce existing knowledge and integrate new knowledge.

Given the minor impact of a few new instances of known classes on the overall data distribution, there's no need to reconstruct existing granular-balls in the knowledge base.
Therefore, these new known class instances are integrated into their corresponding granular-balls, as defined in Definition \ref{known_and_unknown}.
For example, if a known class instance $x_i$ falls within the radius of a granular-ball $GB_j$, it is incorporated into $GB_j$.
This enhances the sample space representation of $GB_j$ and aids in the knowledge consolidation for known classes.
Notably, the addition of $x_i$ to $GB_j$ does not immediately update the granular-ball's center and radius.
Instead, $x_i$ is considered in subsequent calculations for feature redundancy, based on the purity of the granular-ball.
The process of updating a granular ball with a known class instance is defined as follows:

\begin{definition}\label{Adding}(Granular-ball Update for Known Class)
Let $GB$ be a granular-ball in the knowledge base, and let $x$ be a new known class instance whose distance to the center of $GB$ is less than the radius of $GB$.
For $x$, the update for $GB$ is as follows:
\begin{equation}\label{Update_for_Known}
    GB' = GB \cup \{x\}.
\end{equation}
\end{definition}

Another challenge in an open environment is converting unknown class data into known classes and updating the knowledge base accordingly.
Instances that do not belong to any existing granular-ball are deemed unknown class data.
The ideal method for updating the knowledge base involves creating granular-balls for these unknown instances.
However, the absence of supervisory information complicates the identification of specifics, like the exact number and true labels of these unknown classes, preventing the direct construction of label-based granular-balls.

To address this challenge, clustering and setting pseudo-labels are used in this module.
The process begins with clustering the unknown instances and grouping them based on similarities.
Each cluster is then assigned a unique pseudo-label, distinct from existing known class labels.
Utilizing the granular-ball generation method we previously described, granular-balls are created for each cluster.
These new granular-balls are then seamlessly integrated into the existing knowledge base, ensuring its effective update. 
The process for updating granular-balls with unknown class instances is defined as follows:

\begin{definition}\label{Adding}(Granular-ball Update for Unknown Class)
Let $U' = \{x_1, x_2, \dots, x_n\}$ be the new unknown class instances, and let $\mathcal{C} = \{\mathcal{C}_1, \mathcal{C}_2, \dots, \mathcal{C}_n\}$ be the clustering result, and $\mathcal{C}_i$ is a cluster.
$G_{new}$ is a set of granular-balls generated on $\mathcal{C}$ and $G_{ori}$ is a set of existing granular-balls in the knowledge base.
For $U'$, the update of the granular-balls in the knowledge base is as follows:
\begin{equation}\label{Update_for_unKnown}
    G_{all} = G_{ori} \cup G_{new}.
\end{equation}
\end{definition}

The effectiveness of our granular-balls is closely linked to the quality of clustering.
Specifically, the accuracy of granular-ball construction improves when the number of clusters approximates the actual number of distinct data labels.
Consequently, density-based clustering methods are preferred in our framework, as they do not require pre-specifying the number of clusters, unlike methods such as k-means.
In our upcoming experiments, we will utilize DBSCAN \cite{ester1996density}, a density-based clustering algorithm, to demonstrate the efficacy of this approach.

\subsection{Feature Subset Enhancement}
As discussed above, new class instances will constantly emerge in an open and dynamic environment, and class labels play a crucial role in selecting the optimal feature subset for granular-ball neighborhood rough sets.
Since there is no prior research on this problem, it becomes imperative to propose a continuous feature selection method.
Such a method would enable rapid updates to the optimal feature subset and optimize the knowledge base for each period, addressing the dynamic nature of the data.

Continuous feature selection lies in investigating the dynamic pattern of the optimal feature subset before and after the class is increased.
Our motivation analysis and extensive experimental results reveal a key finding: with the addition of new classes, incorporating minimal new features into the existing optimal subset often yields the best results for the current period.
Theoretically, while the previous optimal subset effectively delineates the decision boundary among known classes, the introduction of a new class necessitates additional boundaries for distinction.
Hence, supplementary features are used to represent this additional decision boundary.
\begin{definition}\label{Adding}(Feature Subset Dynamic Pattern)
Given an initial decision table $\langle U_0, A_0, D_0 \rangle$ in period $t_0$ and an unlabeled data stream $U_1$ over $t_1$ consecutive time periods.
Let $S_0$ be the optimal feature subset found on $U_0$.
$L = A-S_0$ is the unselected candidate feature subset in the $t_0$ period.
The optimal feature subset on $U_0 \cup U_1$ in period $t_1$ is defined as:
\begin{equation}\label{optimal_feature_subset_t1}
\begin{gathered}
    S_1 = S_0 \cup B, B \subseteq L\\
    s.t. \quad  GPos_{S_0 \cup B} = GPos_{A}\\
    s.t. \quad  GPos_{S_0 \cup B -\{a\}} \neq GPos_{S_0 \cup B}, a \in B,
\end{gathered}
\end{equation}
\textnormal{where $GPos_{A}$ is the generated positive domain of all features $A$ on data $U_0 \cup U_1$, and $GPos_{S_0 \cup B}$ is the generated positive domain of feature subset ${S_0 \cup B}$.
The combination of conditions $GPos_{S_0 \cup B} = GPos_{A}$ and $GPos_{S_0 \cup B -\{a\}} \neq GPos_{S_0 \cup B}$ entails that $S_0 \cup B$ represents the relative reduction of $A$ as defined in \cite{xia2020gbnrs}.}
\end{definition}

As described in Definition \ref{Adding}, the optimal feature subset $S_1$ of period $t_1$ is obtained based on time period $T_0$.
Specifically, $S_1$ is composed of the optimal subset $S_0$ from $T_0$ and an additional feature $B$ from the candidate features.
Because the generated positive domain can be utilized to stably determine whether a feature should be removed.
Consequently, this allows for the efficient and accurate selection of feature $B$.

When new unknown classes are encountered during the current period, the process of quickly selecting the optimal feature subset can be roughly divided into three steps.
Firstly, based on the optimal feature subset in the knowledge base, the remaining features are considered a candidate subset.
These candidate features, while redundant for known classes, might hold essential information for the new classes.
Next, we evaluate the importance of each candidate feature.
This involves assessing feature redundancy based on the granular-balls in the knowledge base and Eq. \eqref{relative_redundant_feature}.
Specifically, feature $a$ in the candidate subset is deleted, then the granular-balls on $A-\{a\}$ are reconstructed, and the positive domain is calculated by Eq. \eqref{gl_approximation_D}, where $A$ represents all features.
If the positive domain generated undergoes significant changes before and after feature deletion, the feature is deemed important; otherwise, it is considered redundant.
In this way, all features in the candidate subset are examined sequentially.
Finally, the knowledge base is updated to maintain its effectiveness and adaptability to future data.

\subsection{Algorithm Design}
The CFS algorithm consists of two primary stages.
The first stage involves constructing an initial knowledge base using the granular-ball neighborhood rough set.
This process starts with creating a set of granular-balls that cover the entire initial dataset using any granular-ball generation method.
Then, the positive domain is calculated to identify and remove relatively redundant features, resulting in the optimal feature subset for the initial dataset.
Through these findings, the initial knowledge base consisting of constructed granular-balls and optimal feature subsets is obtained.

In the second stage, the algorithm focuses on handling both known and unknown class instances in new data, guided by the existing knowledge base. 
Known class instances are integrated into their corresponding granular-balls, while unknown classes are clustered and assigned pseudo-labels, facilitating their transition from unknown to known.
New granular-balls are then generated for these pseudo-labeled unknown class instances.
These granular-balls are added to the knowledge base, enriching it with new information and thus continuously improving the model's generalization capabilities. 
The final step involves assessing each candidate feature sequentially, using the positive domain generated from the knowledge base.
The changes in this positive domain guide the efficient finding of the optimal feature subset for the current period.

\begin{algorithm2e}[!htb]
\caption{Continual Feature Selection (CFS)}\label{algorithm1}
\LinesNumbered
\SetKwInOut{Input}{Input}\SetKwInOut{Output}{Output}
\Input{Decision table $\langle U_0, A, D \rangle$, unlabeled data stream $U=\langle U_1, U_2, \dots, U_T\rangle$;}
\Output{Selected feature subset $B$;}
\BlankLine
Initialize $B = A$;\\
\If{$t = T_0$}{
// Initial learning \\
    Generate granular-balls $G_0$ of $U_0$ on $B$;\\
    Calculate generation positive domain on $G_0$ by Eq. \eqref{generation_positive_region};\\
    Remove redundant features from $B$ by Eq. \eqref{relative_redundant_feature};\\
    Build initial knowledge base $KB=\{G_0, B\}$ on $U_0$ by Eq. \eqref{knowlwdge_base};\\
}
\Else{
// Open learning \\
    Identify class instances $KC$, $UC$ in $U_t$ by Eq. \eqref{known_class} and Eq. \ref{unknown_class};\\
    Merge $KC$ into $G_0$ by Eq. \eqref{Update_for_Known};\\
    Cluster $UC$ and generate granular-balls $G_{new}$;\\
    Update $KB$ with $G_{new}$ by Eq. \eqref{Update_for_unKnown};\\
    Recalculate generation positive domain on $A$;\\
    \If{$A - B \neq \emptyset$}{
        \For{$a_i \in A - B$}{
            Check redundancy of $a_i$ by Eq. \eqref{relative_redundant_feature};\\
            \If{not redundant}{$B = B \cup \{a_i\}$}
        }
    }
}
\end{algorithm2e}

The CFS algorithm iteratively executes its two-stage process with the arrival of each new data period.
The specifics of this algorithm are detailed in Algorithm \ref{algorithm1}.
Lines 2–8 outline the steps involved in constructing the initial knowledge base using the basic model. 
Lines 10–23 provide a detailed description of the continuous feature selection process that transfers existing knowledge to each new data period.

\section{Experiments}
In this section, we conduct a series of experiments on some real-world datasets to demonstrate the effectiveness and efficiency of the proposed method, especially in terms of the effectiveness of the feature subspace and the efficiency of the feature selection.
Furthermore, the superiority of our method is further verified by taking the results of state-of-the-art feature selection methods on static data as the performance upper bound and comparing with them.

All algorithms were performed with the same software and hardware configuration (CPU: AMD Ryzen 7 4800H @2.90 GHz; RAM: 16 GB; Windows 10; Python 3.8). The source code for the experiments is available at \href{https://github.com/diadai/CFS}{https://github.com/diadai/CFS}.

\subsection{Benchmark Datasets}
We conducted experiments using 15 diverse datasets from the UCI database, including Zoo, Lymphography, Glass, Heart2, Soybean, Anneal, Derm, Vehicle, Segmentation, Pendigits, Dry-Bean, Letter, Sensorless, and Har. 
These datasets, originating from various application domains, differ in the number of classes and features. 
The detailed description of each dataset is summarized in Table \ref{table: dataset}.

\begin{table}[!htb]
\caption{dataset}
\label{table: dataset}
\centering \arraybackslash
\resizebox{0.85\linewidth}{!}{
\begin{tabular}{llllllllll}
\toprule
Id    &Datasets        &Features             &Instance              &Class   \\
    \midrule
1    &Zoo            & 16                    &  101                & 7                 \\
2    &Lymphography   & 18                    &  148                & 4                  \\
3    &Glass          & 9                     &    214              &  7                 \\
4    &Heart2         & 13                    &   303               &  4                 \\
5    &Soybean        & 35                    &  683                & 18                 \\
6    &Anneal         & 38                    &    798              &  5                 \\
7    &Derm           & 34                    &    366              &  6                 \\
8    &Vehicle        & 18                    &    846              &  4                 \\
9    &Segmentation   & 18                    &   2,310             &  7                 \\
10   &Pendigits      & 16                    &  10,992             & 10                 \\
11   &Dry-Bean       & 17                    &  13,611             &  7                 \\
12   &Letter         & 15                    &  20,000             & 16                 \\
13   &Shuttle        & 9                     &  58,000             &  7                 \\
14   &Sensorless     & 48                    &  58,509             & 11                 \\
15   &Har            & 18                    & 165,632             &  5                 \\
\bottomrule
\end{tabular}}
\end{table}

\subsection{Base Model}

Our method is highlighted as a continuous feature selection framework designed for dynamic settings rather than an isolated model. 
It is flexible enough to integrate any GBC method as the underlying model. 
Demonstrating the framework's flexibility, two primary granular-ball models are adopted.

\textbf{GBNRS} \cite{xia2020gbnrs}: As a representative method of GBC, it is also the first method to apply granular-balls to rough sets for feature selection.
It adaptively constructs granular-balls of different sizes and uses their radius as the neighborhood radius, overcoming the parameter adjustment cost caused by the fixed neighborhood radius of the traditional neighborhood rough set.

\textbf{HNRS} \cite{fang2022hypersphere}: As a representative method for ordered granular-ball generation, it is a state-of-the-art neighborhood rough-set model for large datasets.
It maps the original feature space of the data to a high-dimensional feature space through the kernel function and finds the minimum hypersphere for each class, bringing the instances of the same class closer and making the decision boundary clearer.

\subsection{Baselines}
Since this paper is the first to study continual feature selection in the open world, there are no other complete and comparable approaches for this task as the baseline.
Therefore, in order to verify the capability and effectiveness of our proposed framework, we adopt the state-of-the-art static feature selection methods as the baseline.
Meanwhile, for fairness and comprehensiveness, our method will be performed on streaming data.

For each baseline method, its officially released code is reimplemented. 
The details of the baseline methods are listed as follows:

\textbf{Allfeatures}: All original features are employed to be compared with.

\textbf{NSI} \cite{wang2019feature}: It comprehensively takes both upper and lower approximations into the feature evaluation function and proposes neighborhood self-information. 
It is a state-of-the-art algorithm that improves on the classic NRS algorithm in terms of feature selection effectiveness.

\textbf{3WS-RAR} \cite{fang2022three}:
It is a feature selection algorithm suitable for large-scale data presented based on the proposed three-way sampling method and combined with neighborhood rough sets.
It is a state-of-the-art feature selection method that enhances the classic NRS algorithm in large-scale data processing.

\textbf{GIRM} \cite{jing2017quick}: It quantitatively measures the computational cost of various models by inducing multiple indistinguishable relations and proposes a granular structure.
However, unlike the granular-ball method, this granular structure is used to accelerate traditional positive domain calculations.

\subsection{Experiments Settings}
\textbf{Dynamic Settings.}
For the 15 datasets without timestamps, we simulated scenarios in which data classes dynamically change in an open environment. 
Each dataset is randomly divided into initial and new data parts. 
To ensure diversity, these two parts are set based on different proportions of the total number of classes in the dataset. 
The initial part contains either 30\% or 60\% of total classes, adjusted for decimals by rounding down. 
New data streams are set at 10\% or 40\%, rounding up if the product is a decimal.
For instance, the Zoo dataset underwent four dynamic scenarios, pairing different initial and new data ratios: (30\%, 10\%), (30\%, 40\%), (60\%, 10\%), and (60\%, 40\%). 
In the (30\%, 10\%) scenario, we experienced seven periods, each introducing varying class counts: (2, 1, 1, 1, 1, 1, 1). 
This setup started with two classes at period $T_0$ and added one new class in each following period until $T_6$. 
Initially, only the class labels at $T_0$ are known, while the labels from $T_1$ to $T_6$ are unknown.

\textbf{Parameter Settings.}
For experiments involving NSI and 3WS-RAR, the authors' provided source code is employed, with neighborhood radii being finely tuned between (0, 1) by increments of 0.025 for optimal performance. 
We independently implement GIRM's algorithm, adjusting its neighborhood radius from 0.01 to 0.2 at intervals of 0.02, as recommended by the original study for superior outcomes. 
Our framework incorporates the DBSCAN clustering method, setting the neighborhood radius at 0.3 and the minimum core objects at 10. 
The 2-means clustering algorithm is selected for the GBNRS model. 
Gaussian kernel parameters in the HNRS model align with recommendations from \cite{fang2022hypersphere}.
Consistent with insights from \cite{zhang2023incremental}, purities over 0.65 in granular-balls are considered valuable for classification, prompting us to adjust the purity threshold for baseline models from 0.65 to 1 in 0.1 steps.

\textbf{Evaluation Metrics.}
The experiments focus on confirming our framework's effectiveness and efficiency. 
We measure effectiveness through classification accuracy, F1-score, anomaly detection performance, and t-SNE visualizations of the selected feature subsets.
Efficiency is assessed by the average execution time over ten trials on a single device.


\subsection{Effectiveness}


This subsection conducts a comparative analysis of the feature subsets selected by the CFS-HNRS and CFS-GBNRS algorithms in dynamic scenes with the feature subsets selected by the other three methods and the complete feature set in static scenes. 
We optimized the parameters of each dataset via grid search, selecting the highest accuracy for each algorithm's results. 
Three classifiers, kNN, SVM and DT, are used here for verification. 
The experimental results are shown in tables \ref{table: accuracy_knn_hnrs}-\ref{table: accuracy_dt_hnrs}, with the highest values highlighted in bold and asterisks indicating that the results are limited by memory capacity.
The ``$\pm$" symbol denotes the standard deviation of accuracy in ten-fold cross-validation, with higher values indicating more volatility.
Additionally, ``Ave.'' displays the average accuracy of each algorithm across 15 datasets, while ``Opt.'' indicates the frequency with which each algorithm achieves the highest accuracy.

\begin{table*}[!htb]
\caption{Classification accuracy of kNN with different feature selection algorithms.}
\label{table: accuracy_knn_hnrs}
\centering \arraybackslash
\renewcommand\arraystretch{1.1}
\resizebox{\textwidth}{!}{
\setlength{\tabcolsep}{2pt}
\begin{tabular}{@{\extracolsep{\fill}}cccccccccccccccccc}
\toprule
\multirow{2}*{Id}& \multirow{2}*{Allfeatures}& \multirow{2}*{3WS-RAR}   & \multirow{2}*{NSI}   & \multirow{2}*{GIRM}   & \multicolumn{4}{c}{CFS-HNRS}                                                                              & \multicolumn{4}{c}{CFS-GBNRS}                                                               \\
\cmidrule(r){6-9}  \cmidrule(r){10-13}
                    &                     &                          &                            &                          &(30\%,10\%)              &(30\%,40\%)                 &(60\%,10\%)               &(60\%,40\%)              &(30\%,10\%)              &(30\%,40\%)              &(60\%,10\%)         &(60\%,40\%)               \\
\midrule
1                   &$91.00_{\pm 7.00}$   &$92.00_{\pm 8.06}$        &$\pmb{94.00_{\pm 9.17}}$    &$93.00_{\pm 9.00}$        &$92.00_{\pm 7.48}$       &$93.00_{\pm 6.40}$          &$92.00_{\pm 6.00}$        &$93.00_{\pm 6.40}$       &$93.00_{\pm 6.80}$       &$93.00_{\pm 6.60}$       &$91.00_{\pm 7.00}$  &$91.00_{\pm 7.00}$       \\
2                   &$78.57_{\pm 10.59}$  &$73.29_{\pm 12.45}$       &$80.71_{\pm 10.13}$         &$75.71_{\pm 8.57}$        &$77.86_{\pm 11.27}$      &$77.86_{\pm 10.81}$         &$79.29_{\pm 12.96}$       &$79.29_{\pm 12.96}$      &$76.86_{\pm 9.48}$       &$77.14_{\pm 9.48}$       &$78.57_{\pm 10.59}$ &$\pmb{81.43_{\pm 10.48}}$ \\
3                   &$28.10_{\pm 7.51}$   &$29.52_{\pm 5.55}$        &$28.10_{\pm 7.51}$          &$31.38_{\pm 8.19}$        &$30.95_{\pm 9.34}$       &$\pmb{33.81_{\pm 8.10}}$    &$30.48_{\pm 11.11}$       &$31.43_{\pm 10.03}$      &$30.05_{\pm 8.10}$       &$27.67_{\pm 7.44}$       &$31.43_{\pm 5.30}$  &$26.24_{\pm 5.65}$       \\
4                   &$54.33_{\pm 6.84}$   &$57.33_{\pm 8.27}$        &$54.66_{\pm 7.57}$          &$56.33_{\pm 8.09}$        &$58.67_{\pm 7.02}$       &$57.67_{\pm 8.31}$          &$\pmb{59.33_{\pm 5.12}}$  &$59.33_{\pm 8.14}$       &$57.67_{\pm 7.95}$       &$57.67_{\pm 9.32}$       &$59.33_{\pm 8.19}$  &$57.67_{\pm 8.94}$        \\
5                   &$89.41_{\pm 2.61}$   &$84.26_{\pm 4.16}$        &$83.68_{\pm 4.08}$          &$84.12_{\pm 5.33}$        &$89.41_{\pm 2.61}$       &$90.00_{\pm 2.61}$          &$89.41_{\pm 2.61}$        &$89.41_{\pm 2.61}$       &$89.41_{\pm 2.61}$       &$89.58_{\pm 2.87}$       &$89.41_{\pm 2.61}$  &$\pmb{91.54_{\pm 3.48}}$ \\
6                   &$95.32_{\pm 2.27}$   &$96.33_{\pm 2.00}$        &$95.82_{\pm 2.41}$          &$96.20_{\pm 1.96}$        &$95.44_{\pm 1.89}$       &$95.44_{\pm 1.81}$          &$97.47_{\pm 1.79}$        &$\pmb{97.59_{\pm 1.92}}$ &$95.82_{\pm 2.04}$       &$95.32_{\pm 2.27}$       &$92.66_{\pm 2.45}$  &$96.08_{\pm 2.50}$        \\
7                   &$97.50_{\pm 2.31}$   &$96.67_{\pm 1.67}$        &$97.22_{\pm 2.78}$          &$93.33_{\pm 4.84}$        &$97.50_{\pm 2.31}$       &$\pmb{97.50_{\pm 1.94}}$    &$97.50_{\pm 2.62}$        &$\pmb{97.50_{\pm 1.94}}$ &$96.39_{\pm 2.79}$       &$97.50_{\pm 2.31}$       &$97.22_{\pm 2.15}$  &$97.50_{\pm 2.31}$        \\
8                   &$70.48_{\pm 5.32}$   &$70.36_{\pm 5.57}$        &$68.81_{\pm 4.19}$          &$70.60_{\pm 5.14}$        &$\pmb{72.98_{\pm 5.08}}$ &$72.50_{\pm 5.42}$          &$72.50_{\pm 5.42}$        &$72.50_{\pm 5.42}$       &$70.48_{\pm 5.32}$       &$70.48_{\pm 5.32}$       &$70.24_{\pm 5.43}$  &$70.12_{\pm 5.32}$        \\
9                   &$96.15_{\pm 1.09}$   &$96.32_{\pm 0.89}$        &$96.15_{\pm 0.68}$          &$\pmb{96.62_{\pm 1.41}}$  &$96.15_{\pm 1.09}$       &$90.39_{\pm 1.45}$          &$94.89_{\pm 0.96}$        &$91.95_{\pm 2.17}$       &$95.76_{\pm 1.21}$       &$95.93_{\pm 0.87}$       &$96.19_{\pm 0.86}$  &$96.36_{\pm 0.80}$        \\
10                  &$99.34_{\pm 0.24}$   &$99.34_{\pm 0.21}$        &$99.34_{\pm 0.15}$          &$97.82_{\pm 0.42}$        &$99.34_{\pm 0.24}$       &$99.34_{\pm 0.24}$          &$98.90_{\pm 0.29}$        &$99.22_{\pm 0.15}$       &$99.34_{\pm 0.24}$       &$99.34_{\pm 0.16}$       &$99.34_{\pm 0.24}$  &$\pmb{99.37_{\pm 0.22}}$ \\
11                  &$91.54_{\pm 0.63}$   &$91.52_{\pm 0.71}$        &$91.79_{\pm 0.68}$          &$\pmb{91.95_{\pm 0.56}}$  &$91.83_{\pm 0.58}$       &$91.61_{\pm 0.64}$          &$91.83_{\pm 0.58}$        &$91.66_{\pm 0.64}$       &$91.77_{\pm 0.53}$       &$91.90_{\pm 0.58}$       &$91.54_{\pm 0.63}$  &$91.54_{\pm 0.63}$        \\
12                  &$65.62_{\pm 1.05}$   &$64.65_{\pm 0.96}$        &$65.62_{\pm 1.05}$          &$65.62_{\pm 1.05}$        &$\pmb{67.45_{\pm 1.18}}$ &$\pmb{67.45_{\pm 1.18}}$    &$\pmb{67.45_{\pm 1.18}}$  &$66.66_{\pm 0.93}$       &$65.62_{\pm 1.05}$       &$66.02_{\pm 0.87}$       &$65.62_{\pm 1.05}$  &$65.62_{\pm 1.05}$        \\
13                  &$99.90_{\pm 0.04}$   &$99.90_{\pm 0.02}$        &$\ast$                      &$99.90_{\pm 0.04}$        &$99.89_{\pm 0.04}$       &$99.81_{\pm 0.05}$          &$99.91_{\pm 0.03}$        &$99.91_{\pm 0.03}$       &$99.90_{\pm 0.03}$       &$\pmb{99.92_{\pm 0.03}}$ &$99.91_{\pm 0.03}$  &$\pmb{99.92_{\pm 0.03}}$ \\
14                  &$99.02_{\pm 0.09}$   &$99.02_{\pm 0.15}$        &$\ast$                      &$99.30_{\pm 0.10}$        &$99.02_{\pm 0.09}$       &$99.02_{\pm 0.09}$          &$98.94_{\pm 0.09}$        &$98.84_{\pm 0.11}$       &$99.02_{\pm 0.09}$       &$\pmb{99.90_{\pm 0.05}}$ &$99.08_{\pm 0.09}$  &$99.02_{\pm 0.09}$        \\
15                  &$99.49_{\pm 0.04}$   &$99.45_{\pm 0.05}$        &$\ast$                      &$99.36_{\pm 0.05}$        &$99.49_{\pm 0.04}$       &$99.49_{\pm 0.04}$          &$99.49_{\pm 0.04}$        &$99.49_{\pm 0.04}$       &$\pmb{99.50_{\pm 0.04}}$ &$99.49_{\pm 0.05}$       &$99.49_{\pm 0.04}$  &$99.49_{\pm 0.04}$        \\
\midrule
Ave.                &$83.72_{\pm 3.18}$   &$83.33_{\pm 3.38}$        &$79.66_{\pm 4.20}$          &$83.46_{\pm 3.65}$        &$84.53_{\pm 3.35}$       &$84.33_{\pm 3.27}$          &$\pmb{84.62_{\pm 3.39}}$  &$84.52_{\pm 3.57}$       &$84.04_{\pm 3.22}$       &$84.06_{\pm 3.21}$       &$84.07_{\pm 3.11}$  &$84.19_{\pm 3.24}$  \\
Opt.                &0                    &0                         &1                           &2                         &2                        &3                           &2                         &2                        &1                        &2                        &0                   &$\pmb{4}$             \\
\bottomrule
\end{tabular}}
\end{table*}

Tables \ref{table: accuracy_knn_hnrs}-\ref{table: accuracy_dt_hnrs} list the classification accuracies of various comparison algorithms on the three classifiers.
Notably, the subset of generated features should exhibit either unchanged or improved classification accuracy on the classifier compared to using all features.
The tabular data demonstrates that, in the majority of cases, the subsets selected by our method not only valid feature subsets of the original features but also exhibit superior accuracy compared to other feature selection methods.
One of the reasons is that the basic model of our framework is the granular-ball method, which adaptively granulates data to generate appropriate neighborhood radii, thereby circumventing the heterogeneous transmission phenomena caused by fixed neighborhood radii.
Additionally, the adoption of coarser granularity through granular-ball models enhances robustness against noise points or outliers in proximity to decision boundaries.
Thus, in comparison to conventional NRS approaches, our CFS-HNRS and CFS-GBNRS exhibit markedly improved performance.

\begin{table*}[!htb]
\caption{Classification accuracy of SVM with different feature selection algorithms.}
\label{table: accuracy_svm_hnrs}
\centering \arraybackslash
\renewcommand\arraystretch{1.1}
\resizebox{\textwidth}{!}{
\setlength{\tabcolsep}{2pt}
\begin{tabular}{@{\extracolsep{\fill}}cccccccccccccccccccc}
\toprule
\multirow{2}*{Id}& \multirow{2}*{Allfeatures}& \multirow{2}*{3WS-RAR}   & \multirow{2}*{NSI}   & \multirow{2}*{GIRM}   & \multicolumn{4}{c}{CFS-HNRS}                                                                        & \multicolumn{4}{c}{CFS-GBNRS}                                                               \\
\cmidrule(r){6-9}  \cmidrule(r){10-13}
                    &                     &                          &                         &                         &(30\%,10\%)              &(30\%,40\%)              &(60\%,10\%)              &(60\%,40\%)                &(30\%,10\%)                &(30\%,40\%)            &(60\%,10\%)             &(60\%,40\%)              \\
\midrule
1                   &$94.00_{\pm 6.63}$   &$93.33_{\pm 11.00}$       &$93.00_{\pm 9.43}$       &$92.00_{\pm 7.48}$       &$93.00_{\pm 6.40}$       &$95.00_{\pm 6.71}$       &$94.00_{\pm 6.63}$       &$95.00_{\pm 6.71}$       &$\pmb{96.00_{\pm 6.71}}$   &$92.00_{\pm 11.83}$      &$94.00_{\pm 6.63}$       &$94.00_{\pm 6.63}$       \\
2                   &$83.57_{\pm 11.54}$  &$77.57_{\pm 11.07}$       &$\pmb{84.29_{\pm 7.69}}$ &$77.14_{\pm 10.50}$      &$77.14_{\pm 13.09}$      &$80.71_{\pm 12.39}$      &$80.71_{\pm 12.39}$      &$80.71_{\pm 12.39}$      &$83.57_{\pm 11.54}$        &$83.72_{\pm 11.79}$      &$83.57_{\pm 11.54}$      &$83.29_{\pm 12.79}$       \\
3                   &$36.19_{\pm 12.81}$  &$32.10_{\pm 8.10}$        &$37.29_{\pm 11.63}$      &$36.81_{\pm 8.90}$       &$35.71_{\pm 11.71}$      &$36.67_{\pm 11.48}$      &$\pmb{39.52_{\pm 12.79}}$&$37.14_{\pm 12.92}$      &$38.24_{\pm 8.83}$         &$38.71_{\pm 9.58}$       &$35.90_{\pm 10.22}$      &$36.76_{\pm 10.74}$      \\
4                   &$55.33_{\pm 8.06}$   &$57.33_{\pm 7.12}$        &$55.99_{\pm 9.71}$       &$57.00_{\pm 11.53}$      &$59.00_{\pm 6.84}$       &$59.33_{\pm 7.27}$       &$59.33_{\pm 5.12}$       &$\pmb{60.33_{\pm 8.23}}$ &$56.34_{\pm 10.23}$        &$58.33_{\pm 9.52}$       &$\pmb{60.33_{\pm 8.23}}$ &$57.00_{\pm 9.29}$       \\
5                   &$93.97_{\pm 2.50}$   &$86.32_{\pm 2.55}$        &$90.44_{\pm 2.89}$       &$92.65_{\pm 3.42}$       &$\pmb{93.97_{\pm 2.50}}$ &$93.53_{\pm 2.56}$       &$\pmb{93.97_{\pm 2.50}}$ &$\pmb{93.97_{\pm 2.50}}$ &$\pmb{93.97_{\pm 2.50}}$   &$93.09_{\pm 3.29}$       &$\pmb{93.97_{\pm 2.50}}$ &$93.09_{\pm 2.79}$       \\
6                   &$94.43_{\pm 2.48}$   &$92.28_{\pm 2.86}$        &$93.42_{\pm 2.18}$       &$90.76_{\pm 3.05}$       &$93.67_{\pm 3.58}$       &$94.43_{\pm 2.48}$       &$\pmb{95.82_{\pm 2.66}}$ &$95.57_{\pm 2.73}$       &$95.19_{\pm 2.58}$         &$94.43_{\pm 2.48}$       &$93.67_{\pm 2.45}$       &$94.30_{\pm 2.79}$       \\
7                   &$98.33_{\pm 1.36}$   &$97.50_{\pm 2.31}$        &$\pmb{98.33_{\pm 1.36}}$ &$94.17_{\pm 3.82}$       &$\pmb{98.33_{\pm 1.36}}$ &$98.06_{\pm 1.78}$       &$\pmb{98.33_{\pm 1.84}}$ &$\pmb{98.33_{\pm 1.84}}$ &$97.22_{\pm 1.24}$         &$\pmb{98.33_{\pm 1.36}}$ &$\pmb{98.33_{\pm 1.36}}$ &$\pmb{98.33_{\pm 1.36}}$ \\
8                   &$75.95_{\pm 3.40}$   &$75.48_{\pm 4.27}$        &$72.86_{\pm 3.79}$       &$74.76_{\pm 4.54}$       &$75.83_{\pm 3.88}$       &$75.12_{\pm 3.51}$       &$\pmb{76.19_{\pm 3.98}}$ &$\pmb{76.19_{\pm 3.98}}$ &$75.95_{\pm 3.40}$         &$75.95_{\pm 3.40}$       &$75.60_{\pm 3.89}$       &$72.26_{\pm 4.36}$       \\
9                   &$93.94_{\pm 1.88}$   &$\pmb{94.55_{\pm 1.65}}$  &$94.42_{\pm 1.55}$       &$93.81_{\pm 1.49}$       &$93.94_{\pm 1.88}$       &$87.14_{\pm 2.42}$       &$93.12_{\pm 1.58}$       &$87.62_{\pm 2.40}$       &$94.07_{\pm 1.46}$         &$94.16_{\pm 1.59}$       &$94.20_{\pm 1.77}$       &$94.20_{\pm 1.49}$       \\
10                  &$99.44_{\pm 0.17}$   &$99.43_{\pm 0.12}$        &$\pmb{99.44_{\pm 0.17}}$ &$97.56_{\pm 0.40}$       &$\pmb{99.44_{\pm 0.17}}$ &$\pmb{99.44_{\pm 0.17}}$ &$99.43_{\pm 0.15}$       &$99.14_{\pm 0.14}$       &$\pmb{99.44_{\pm 0.17}}$   &$99.32_{\pm 0.20}$       &$\pmb{99.44_{\pm 0.17}}$ &$99.44_{\pm 0.24}$       \\
11                  &$92.42_{\pm 0.60}$   &$92.67_{\pm 0.75}$        &$92.46_{\pm 0.64}$       &$\pmb{92.67_{\pm 0.55}}$ &$92.42_{\pm 0.60}$       &$92.43_{\pm 0.58}$       &$92.42_{\pm 0.60}$       &$92.37_{\pm 0.60}$       &$92.59_{\pm 0.60}$         &$\pmb{92.67_{\pm 0.61}}$ &$92.42_{\pm 0.60}$       &$92.42_{\pm 0.60}$       \\
12                  &$58.16_{\pm 1.51}$   &$56.62_{\pm 0.70}$        &$58.16_{\pm 1.51}$       &$58.16_{\pm 1.51}$       &$58.16_{\pm 1.51}$       &$58.16_{\pm 1.51}$       &$58.16_{\pm 1.51}$       &$\pmb{58.63_{\pm 1.41}}$ &$58.16_{\pm 1.51}$         &$57.41_{\pm 0.63}$       &$58.16_{\pm 1.51}$       &$58.16_{\pm 1.51}$       \\
13                  &$99.64_{\pm 0.05}$   &$99.61_{\pm 0.10}$        &$\ast$                   &$99.64_{\pm 0.05}$       &$99.64_{\pm 0.05}$       &$99.62_{\pm 0.05}$       &$99.64_{\pm 0.05}$       &$99.56_{\pm 0.07}$       &$\pmb{99.65_{\pm 0.06}}$  &$99.63_{\pm 0.06}$        &$\pmb{99.65_{\pm 0.06}}$ &$99.64_{\pm 0.06}$       \\
14                  &$90.41_{\pm 0.52}$   &$90.38_{\pm 0.55}$        &$\ast$                   &$92.08_{\pm 0.51}$       &$90.41_{\pm 0.52}$       &$90.41_{\pm 0.52}$       &$90.30_{\pm 0.49}$       &$90.18_{\pm 0.45}$       &$90.41_{\pm 0.52}$         &$\pmb{92.13_{\pm 0.43}}$ &$90.46_{\pm 0.52}$       &$90.41_{\pm 0.52}$       \\
15                  &$91.64_{\pm 0.18}$   &$\pmb{93.75_{\pm 0.18}}$  &$\ast$                   &$91.28_{\pm 0.22}$       &$91.64_{\pm 0.18}$       &$91.64_{\pm 0.18}$       &$91.64_{\pm 0.18}$       &$92.83_{\pm 0.18}$       &$92.88_{\pm 0.14}$         &$92.77_{\pm 0.11}$       &$91.90_{\pm 0.21}$       &$92.20_{\pm 0.17}$       \\
\midrule
Ave.               &$83.83_{\pm 3.58}$   &$82.59_{\pm 3.56}$        &$80.84_{\pm 4.32}$       &$82.70_{\pm 3.86}$       &$83.49_{\pm 3.62}$       &$83.45_{\pm 3.57}$       &$84.17_{\pm 3.50}$       &$83.84_{\pm 3.77}$       &$\pmb{84.25_{\pm 3.43}}$    &$84.18_{\pm 3.79}$       &$84.11_{\pm 3.54}$       &$83.70_{\pm 3.69}$       \\
Opt.               &0                    &2                         &3                        &1                        &3                        &1                        &$\pmb{5}$                &$\pmb{5}$                &4                           &3                        &$\pmb{5}$                &1                        \\

\bottomrule
\end{tabular}}
\end{table*}

\begin{table*}[!htb]
\caption{Classification accuracy of DT with different feature selection algorithms.}
\label{table: accuracy_dt_hnrs}
\centering \arraybackslash
\renewcommand\arraystretch{1.1}
\resizebox{\textwidth}{!}{
\setlength{\tabcolsep}{2pt}
\begin{tabular}{@{\extracolsep{\fill}}ccccccccccccccccccccc}
\toprule
\multirow{2}*{Id}& \multirow{2}*{Allfeatures}& \multirow{2}*{3WS-RAR}   & \multirow{2}*{NSI}   & \multirow{2}*{GIRM}   & \multicolumn{4}{c}{CFS-HNRS}                                                                        & \multicolumn{4}{c}{CFS-GBNRS}                                                               \\
\cmidrule(r){6-9}  \cmidrule(r){10-13}
                    &                     &                          &                     &                          &(30\%,10\%)               &(30\%,40\%)                 &(60\%,10\%)               &(60\%,40\%)               &(30\%,10\%)        &(30\%,40\%)              &(60\%,10\%)         &(60\%,40\%)             \\
\midrule
1                   &$95.00_{\pm 6.71}$   &$96.00_{\pm 6.63}$        &$97.00_{\pm 3.00}$   &$\pmb{98.00_{\pm 4.00}}$  &$95.00_{\pm 6.71}$        &$97.00_{\pm 4.58}$        &$95.00_{\pm 5.00}$       &$96.00_{\pm 6.63}$       &$96.00_{\pm 4.58}$ &$95.00_{\pm 6.71}$       &$95.00_{\pm 6.71}$  &$95.00_{\pm 6.71}$      \\
2                   &$76.43_{\pm 10.62}$  &$73.57_{\pm 10.62}$       &$76.29_{\pm 9.15}$   &$78.71_{\pm 10.13}$       &$79.29_{\pm 8.11}$        &$76.43_{\pm 9.61}$        &$82.86_{\pm 9.69}$       &$\pmb{83.57_{\pm 9.61}}$ &$76.43_{\pm 10.62}$&$76.14_{\pm 9.62}$       &$76.43_{\pm 10.62}$ &$80.43_{\pm 10.61}$      \\
3                   &$22.86_{\pm 6.32}$   &$22.24_{\pm 8.53}$        &$21.81_{\pm 7.68}$   &$21.90_{\pm 9.81}$        &$28.10_{\pm 13.21}$       &$\pmb{32.86_{\pm 12.68}}$ &$29.52_{\pm 8.46}$       &$28.10_{\pm 7.51}$       &$26.67_{\pm 8.02}$ &$26.67_{\pm 8.02}$       &$26.00_{\pm 9.29}$  &$27.62_{\pm 12.92}$     \\
4                   &$48.33_{\pm 12.41}$  &$48.33_{\pm 9.91}$        &$46.67_{\pm 9.55}$   &$50.33_{\pm 7.67}$        &$51.33_{\pm 6.18}$        &$53.33_{\pm 6.15}$        &$50.00_{\pm 6.67}$       &$51.33_{\pm 5.62}$       &$51.00_{\pm 8.03}$ &$51.33_{\pm 8.36}$       &$54.00_{\pm 9.98}$  &$\pmb{55.67_{\pm 9.45}}$\\
5                   &$91.47_{\pm 2.61}$   &$90.09_{\pm 2.97}$        &$88.94_{\pm 4.73}$   &$92.94_{\pm 3.82}$        &$91.47_{\pm 2.61}$        &$92.06_{\pm 2.65}$        &$91.47_{\pm 2.61}$       &$91.47_{\pm 2.61}$       &$91.47_{\pm 2.61}$ &$\pmb{92.35_{\pm 2.93}}$ &$91.47_{\pm 2.61}$  &$92.21_{\pm 2.19}$      \\
6                   &$99.11_{\pm 1.17}$   &$98.48_{\pm 1.48}$        &$98.29_{\pm 2.73}$   &$98.35_{\pm 1.39}$        &$97.22_{\pm 2.25}$        &$99.24_{\pm 1.01}$        &$\pmb{99.49_{\pm 0.84}}$ &$\pmb{99.49_{\pm 0.84}}$ &$99.11_{\pm 1.50}$ &$99.37_{\pm 0.85}$       &$98.00_{\pm 1.58}$  &$99.24_{\pm 1.16}$      \\
7                   &$93.33_{\pm 3.33}$   &$94.56_{\pm 2.55}$        &$91.78_{\pm 1.84}$   &$93.89_{\pm 3.89}$        &$\pmb{94.72_{\pm 2.15}}$  &$94.17_{\pm 3.15}$        &$93.89_{\pm 2.08}$       &$93.89_{\pm 2.72}$       &$93.06_{\pm 3.11}$ &$93.06_{\pm 3.57}$       &$92.78_{\pm 3.33}$  &$93.06_{\pm 3.57}$      \\
8                   &$71.07_{\pm 3.53}$   &$68.81_{\pm 3.60}$        &$70.60_{\pm 5.24}$   &$69.05_{\pm 4.61}$        &$71.90_{\pm 3.77}$        &$\pmb{72.50_{\pm 4.04}}$  &$71.67_{\pm 3.98}$       &$72.14_{\pm 2.88}$       &$71.07_{\pm 3.53}$ &$71.07_{\pm 3.53}$       &$69.86_{\pm 5.30}$  &$68.67_{\pm 5.63}$      \\
9                   &$96.41_{\pm 1.37}$   &$95.80_{\pm 1.08}$        &$96.61_{\pm 1.06}$   &$96.10_{\pm 1.34}$        &$96.41_{\pm 1.37}$        &$91.26_{\pm 1.83}$        &$96.23_{\pm 0.97}$       &$91.69_{\pm 2.26}$       &$96.58_{\pm 1.26}$ &$96.36_{\pm 1.08}$       &$96.80_{\pm 0.65}$  &$\pmb{97.10_{\pm 1.11}}$\\
10                  &$96.33_{\pm 0.58}$   &$96.45_{\pm 0.51}$        &$96.01_{\pm 0.82}$   &$94.36_{\pm 0.67}$        &$96.33_{\pm 0.58}$        &$96.33_{\pm 0.58}$        &$96.11_{\pm 0.59}$       &$96.11_{\pm 0.71}$       &$96.33_{\pm 0.58}$ &$96.27_{\pm 0.68}$       &$96.33_{\pm 0.58}$  &$\pmb{96.54_{\pm 0.29}}$\\
11                  &$89.65_{\pm 0.79}$   &$89.40_{\pm 0.82}$        &$89.77_{\pm 0.80}$   &$89.91_{\pm 0.95}$        &$89.65_{\pm 0.79}$        &$89.94_{\pm 0.83}$        &$89.65_{\pm 0.79}$       &$89.13_{\pm 0.86}$       &$89.88_{\pm 0.76}$ &$\pmb{89.99_{\pm 0.63}}$ &$89.65_{\pm 0.79}$  &$89.65_{\pm 0.79}$      \\
12                  &$62.62_{\pm 1.23}$   &$61.61_{\pm 0.97}$        &$62.62_{\pm 1.23}$   &$62.62_{\pm 1.23}$        &$62.62_{\pm 1.23}$        &$62.62_{\pm 1.23}$        &$62.62_{\pm 1.23}$       &$63.08_{\pm 1.00}$       &$62.62_{\pm 1.23}$ &$\pmb{63.84_{\pm 1.10}}$ &$62.62_{\pm 1.23}$  &$62.62_{\pm 1.23}$      \\
13                  &$99.99_{\pm 0.01}$   &$99.99_{\pm 0.02}$        &$\ast$               &$\pmb{99.99_{\pm 0.01}}$  &$99.97_{\pm 0.03}$        &$99.84_{\pm 0.05}$        &$99.98_{\pm 0.02}$       &$99.98_{\pm 0.02}$       &$99.98_{\pm 0.02}$ &$99.98_{\pm 0.02}$       &$99.98_{\pm 0.02}$  &$99.98_{\pm 0.02}$      \\
14                  &$98.49_{\pm 0.16}$   &$98.61_{\pm 0.18}$        &$\ast$               &$98.56_{\pm 0.10}$        &$98.49_{\pm 0.16}$        &$98.49_{\pm 0.16}$        &$98.52_{\pm 0.10}$       &$98.49_{\pm 0.18}$       &$98.49_{\pm 0.16}$ &$\pmb{98.77_{\pm 0.15}}$ &$98.65_{\pm 0.17}$  &$98.49_{\pm 0.16}$      \\
15                  &$98.48_{\pm 0.13}$   &$98.42_{\pm 0.07}$        &$\ast$               &$98.12_{\pm 0.11}$        &$98.48_{\pm 0.13}$        &$98.48_{\pm 0.13}$        &$98.48_{\pm 0.13}$       &$\pmb{98.49_{\pm 0.14}}$ &$98.45_{\pm 0.10}$ &$98.43_{\pm 0.09}$       &$98.47_{\pm 0.09}$  &$\pmb{98.49_{\pm 0.09}}$\\
\midrule
Ave.                &$82.64_{\pm 3.40}$   &$82.16_{\pm 3.33}$        &$78.03_{\pm 3.99}$   &$82.86_{\pm 3.32}$        &$83.38_{\pm 3.29}$        &$83.64_{\pm 3.25}$        &$\pmb{83.70_{\pm 2.88}}$ &$83.53_{\pm 2.91}$       &$83.14_{\pm 3.07}$ &$83.24_{\pm 3.16}$       &$83.07_{\pm 3.53}$  &$83.65_{\pm 3.73}$        \\
Opt.                &0                    &0                         &0                    &2                         &1                         &2                         &1                        &3                        &0                  &$\pmb{4}$                &0                   &$\pmb{4}$                 \\

\bottomrule
\end{tabular}}
\end{table*}

\begin{figure*}[!htb]
\centering
\includegraphics[width=\linewidth]{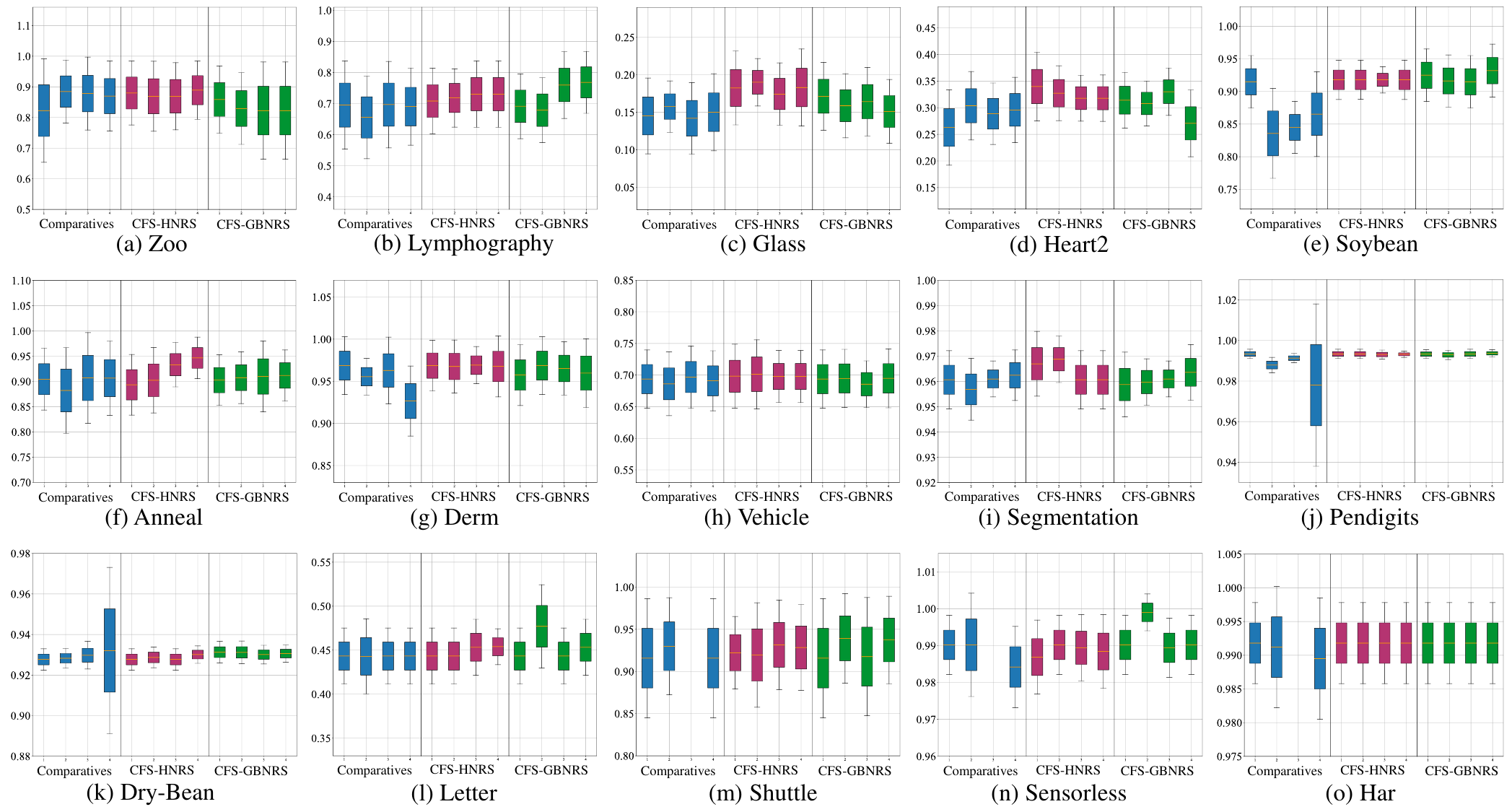}
\caption{Comparison of different feature selection algorithms on classification F1-score of kNN. The y-axis indicates the F-score value, while the x-axis displays the comparison methods in order: Allfeatures, 3WS-RAR, NSI, GIRM, followed by our two methods under four dynamic settings: (30\%, 10\%), (30\%, 40\%), (60\%, 10\%), and (60\%, 40\%), moving left to right.}
\label{fig: F1}
\end{figure*}

\begin{figure}[!htb]
\centering
\includegraphics[width=0.92\linewidth]{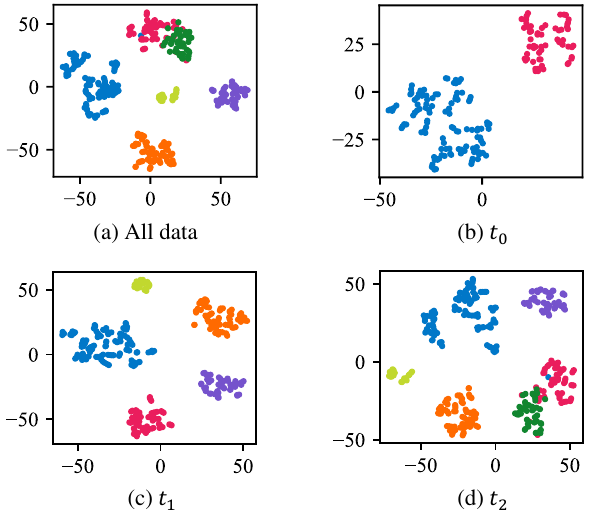}
\caption{Visualization of t-SNE on the Derm dataset. Each color in the t-SNE plot represents a class. (a) illustrates the original data distribution, while figures (b), (c), and (d) depict the data distributions for the optimal feature subsets across various periods in the data stream.}
\label{fig: t-SNE}
\end{figure}


The analysis of the CFS-HNRS and CFS-GBNRS methods, as shown in tables \ref{table: accuracy_knn_hnrs} to \ref{table: accuracy_dt_hnrs}, suggests that better outcomes are achieved in the open learning phase when more data is added in each period, given the same initial learning stage data. 
This improvement may be attributed to the fact that dynamically adding data in small proportions can result in certain periods where only one class of data is added, which is then clustered into the same pseudo-label cluster by DBSCAN.
Then, this leads to an early stopping of the granular ball generation process due to easily reaching the purity threshold, resulting in overly large granular balls. 
Ultimately, the dynamically generated positive domain in the feature subset dynamic pattern is more prone to change, leading to a feature subset that is closer to the original features but not truly lean enough. 
Moreover, in most instances, when the increase ratio is the same, experimental outcomes are marginally superior with 60\% of the initial data compared to 30\%.
Fortunately, both different initial data settings result in efficient subsets of features in comparison to all features.
This indicates that our method does not overly rely on the volume of initial data, demonstrating its generalizability. 
Therefore, in practical applications, it is recommended to extend the data collection period to increase the volume of data for a new period, thereby achieving better results.


To avoid the impact of instance imbalance on classification accuracy in some datasets, we also compared the classification F1-score of different methods on kNN.
Box plots are used to visually analyze the experimental results, as shown in Fig. \ref{fig: F1}.
The comparative method is denoted in blue, whereas our CFS-HNRS and CFS-GBNRS methods are illustrated in red and green, respectively. 
The yellow line in the box plot represents the median of ten F1-scores, while the upper and lower limits indicate the maximum and minimum values.
The box's length signifies the F1-scores' variance, where longer boxes suggest more significant fluctuations. 
Intuitively, our methods surpass the comparison methods in F1-score on the majority of datasets and demonstrate better stability.

In addition to classification accuracy, we assessed the efficacy of our method through the lens of four anomaly detection methods, aiming to validate the utility of the feature subset generated by our approach in an open environment for anomaly detection. 
The experimental results are shown in Table \ref{table: detection_accuracy}.
The table shows that in most cases, the four anomaly detection methods maintain or improve the original anomaly detection accuracy on the feature subsets we selected.
Specifically, our method works best on One-Class SVM, improving the original anomaly detection accuracy on multiple datasets.
In the Isolation Forest method, we maintain the same accuracy as the original features.
For the Local Outlier Factor method, the accuracy has slightly decreased on some datasets, yet this reduction was kept within a 0.07 range. 
Overall, these results substantiate the effectiveness of our method.

\begin{table*}[!htb]
\caption{Detection accuracy of different anomaly detection algorithms before and after feature selection.}
\label{table: detection_accuracy}
\centering \arraybackslash
\renewcommand\arraystretch{1.2}
\resizebox{1\textwidth}{!}{
\setlength{\tabcolsep}{2pt}
\begin{tabular}{@{\extracolsep{\fill}}cccccccccccccccccccccc}
\toprule
\multirow{2}*{Id}   &\multicolumn{3}{c}{Robust Covariance}                                       & \multicolumn{3}{c}{One-Class SVM}                        &\multicolumn{3}{c}{Isolation Forest}                       & \multicolumn{3}{c}{Local Outlier Factor}                                                                                \\
\cmidrule(r){2-4}  \cmidrule(r){5-7}    \cmidrule(r){8-10}     \cmidrule(r){11-13}
                    & Allfeatures              & CFS-HNRS             & CFS-GBNRS           & Allfeatures           & CFS-HNRS          & CFS-GBNRS         & Allfeatures         & CFS-HNRS         & CFS-GBNRS             & Allfeatures        & CFS-HNRS          & CFS-GBNRS           \\
\midrule
1                   &$1.00_{\pm 0.00}$      &$1.00_{\pm 0.00}$     &$1.00_{\pm 0.00}$    &$0.86_{\pm 0.01}$   &$0.90_{\pm 0.01}$  &$0.88_{\pm 0.04}$  &$1.00_{\pm 0.00}$&$1.00_{\pm 0.00}$  &$1.00_{\pm 0.00}$      &$1.00_{\pm 0.00}$&$1.00_{\pm 0.00}$  &$1.00_{\pm 0.00}$     \\
2                   &$1.00_{\pm 0.00}$      &$1.00_{\pm 0.00}$     &$1.00_{\pm 0.00}$    &$0.90_{\pm 0.00}$   &$0.91_{\pm 0.02}$  &$0.95_{\pm 0.02}$  &$1.00_{\pm 0.00}$&$1.00_{\pm 0.00}$  &$1.00_{\pm 0.00}$      &$1.00_{\pm 0.00}$&$1.00_{\pm 0.00}$  &$1.00_{\pm 0.00}$     \\
3                   &$0.99_{\pm 0.00}$      &$0.99_{\pm 0.00}$     &$0.99_{\pm 0.00}$    &$0.96_{\pm 0.02}$   &$0.97_{\pm 0.01}$  &$0.98_{\pm 0.01}$  &$1.00_{\pm 0.00}$&$1.00_{\pm 0.00}$  &$1.00_{\pm 0.00}$      &$1.00_{\pm 0.00}$&$1.00_{\pm 0.00}$  &$1.00_{\pm 0.00}$     \\
4                   &$1.00_{\pm 0.00}$      &$1.00_{\pm 0.00}$     &$1.00_{\pm 0.00}$    &$0.95_{\pm 0.01}$   &$0.99_{\pm 0.01}$  &$0.95_{\pm 0.01}$  &$1.00_{\pm 0.00}$&$1.00_{\pm 0.00}$  &$1.00_{\pm 0.00}$      &$1.00_{\pm 0.00}$&$1.00_{\pm 0.00}$  &$1.00_{\pm 0.00}$     \\
5                   &$1.00_{\pm 0.00}$      &$1.00_{\pm 0.00}$     &$1.00_{\pm 0.00}$    &$0.92_{\pm 0.00}$   &$0.94_{\pm 0.00}$  &$0.92_{\pm 0.13}$  &$1.00_{\pm 0.00}$&$1.00_{\pm 0.00}$  &$1.00_{\pm 0.00}$      &$1.00_{\pm 0.00}$&$1.00_{\pm 0.00}$  &$0.95_{\pm 0.01}$     \\
6                   &$1.00_{\pm 0.00}$      &$1.00_{\pm 0.00}$     &$1.00_{\pm 0.00}$    &$0.97_{\pm 0.00}$   &$0.98_{\pm 0.00}$  &$0.98_{\pm 0.00}$  &$1.00_{\pm 0.00}$&$1.00_{\pm 0.00}$  &$1.00_{\pm 0.00}$      &$1.00_{\pm 0.00}$&$1.00_{\pm 0.00}$  &$1.00_{\pm 0.00}$     \\
7                   &$1.00_{\pm 0.00}$      &$1.00_{\pm 0.00}$     &$1.00_{\pm 0.00}$    &$0.94_{\pm 0.00}$   &$0.98_{\pm 0.00}$  &$0.92_{\pm 0.00}$  &$1.00_{\pm 0.00}$&$1.00_{\pm 0.00}$  &$1.00_{\pm 0.00}$      &$1.00_{\pm 0.00}$&$1.00_{\pm 0.00}$  &$1.00_{\pm 0.00}$     \\
8                   &$1.00_{\pm 0.00}$      &$1.00_{\pm 0.00}$     &$1.00_{\pm 0.00}$    &$0.97_{\pm 0.00}$   &$1.00_{\pm 0.00}$  &$1.00_{\pm 0.00}$  &$1.00_{\pm 0.00}$&$1.00_{\pm 0.00}$  &$1.00_{\pm 0.00}$      &$1.00_{\pm 0.00}$&$1.00_{\pm 0.00}$  &$1.00_{\pm 0.00}$     \\
9                   &$1.00_{\pm 0.00}$      &$1.00_{\pm 0.00}$     &$0.99_{\pm 0.00}$    &$1.00_{\pm 0.00}$   &$1.00_{\pm 0.00}$  &$1.00_{\pm 0.00}$  &$1.00_{\pm 0.00}$&$1.00_{\pm 0.00}$  &$1.00_{\pm 0.00}$      &$1.00_{\pm 0.00}$&$1.00_{\pm 0.00}$  &$1.00_{\pm 0.00}$     \\
10                  &$1.00_{\pm 0.00}$      &$1.00_{\pm 0.00}$     &$1.00_{\pm 0.00}$    &$1.00_{\pm 0.00}$   &$1.00_{\pm 0.00}$  &$1.00_{\pm 0.00}$  &$1.00_{\pm 0.00}$&$1.00_{\pm 0.00}$  &$1.00_{\pm 0.00}$      &$1.00_{\pm 0.00}$&$1.00_{\pm 0.00}$  &$1.00_{\pm 0.00}$     \\
11                  &$1.00_{\pm 0.00}$      &$1.00_{\pm 0.00}$     &$1.00_{\pm 0.00}$    &$1.00_{\pm 0.00}$   &$1.00_{\pm 0.00}$  &$1.00_{\pm 0.00}$  &$1.00_{\pm 0.00}$&$1.00_{\pm 0.00}$  &$1.00_{\pm 0.00}$      &$0.99_{\pm 0.00}$&$0.97_{\pm 0.00}$  &$0.93_{\pm 0.00}$     \\
12                  &$1.00_{\pm 0.00}$      &$1.00_{\pm 0.00}$     &$1.00_{\pm 0.00}$    &$1.00_{\pm 0.00}$   &$1.00_{\pm 0.00}$  &$1.00_{\pm 0.00}$  &$1.00_{\pm 0.00}$&$1.00_{\pm 0.00}$  &$1.00_{\pm 0.00}$      &$0.98_{\pm 0.00}$&$0.97_{\pm 0.00}$  &$0.92_{\pm 0.00}$     \\
13                  &$1.00_{\pm 0.00}$      &$1.00_{\pm 0.00}$     &$1.00_{\pm 0.00}$    &$1.00_{\pm 0.00}$   &$1.00_{\pm 0.00}$  &$1.00_{\pm 0.00}$  &$1.00_{\pm 0.00}$&$1.00_{\pm 0.00}$  &$1.00_{\pm 0.00}$      &$0.90_{\pm 0.00}$&$0.90_{\pm 0.00}$  &$0.90_{\pm 0.00}$     \\
14                  &$1.00_{\pm 0.00}$      &$1.00_{\pm 0.00}$     &$1.00_{\pm 0.00}$    &$1.00_{\pm 0.00}$   &$1.00_{\pm 0.00}$  &$1.00_{\pm 0.00}$  &$1.00_{\pm 0.00}$&$1.00_{\pm 0.00}$  &$1.00_{\pm 0.00}$      &$1.00_{\pm 0.00}$&$1.00_{\pm 0.00}$  &$0.93_{\pm 0.00}$     \\
15                  &$1.00_{\pm 0.00}$      &$1.00_{\pm 0.00}$     &$1.00_{\pm 0.00}$    &$1.00_{\pm 0.00}$   &$1.00_{\pm 0.00}$  &$1.00_{\pm 0.00}$  &$1.00_{\pm 0.00}$&$1.00_{\pm 0.00}$  &$1.00_{\pm 0.00}$      &$0.91_{\pm 0.00}$&$0.91_{\pm 0.00}$  &$0.91_{\pm 0.00}$     \\

\bottomrule

\end{tabular}}
\end{table*}

As shown in Fig. \ref{fig: t-SNE}, we visualized the data distribution to intuitively assess the effectiveness of the selected feature subsets based on decision boundaries.
The class dynamics for the Derm dataset were set to (30\%, 40\%), corresponding to (2, 3, 1).
t-SNE visualizations also indicate that our method maintains clear decision boundaries for each class across periods, despite the continuous increase of unknown classes, exemplifying the ideal behavior for this problem.
This suggests that our framework, despite its simplicity, effectively agglomerates samples of each class in the feature space, thereby validating the effectiveness of the chosen feature subsets.

\subsection{Efficiency}

\begin{figure*}[h]
\centering
\includegraphics[width=\linewidth]{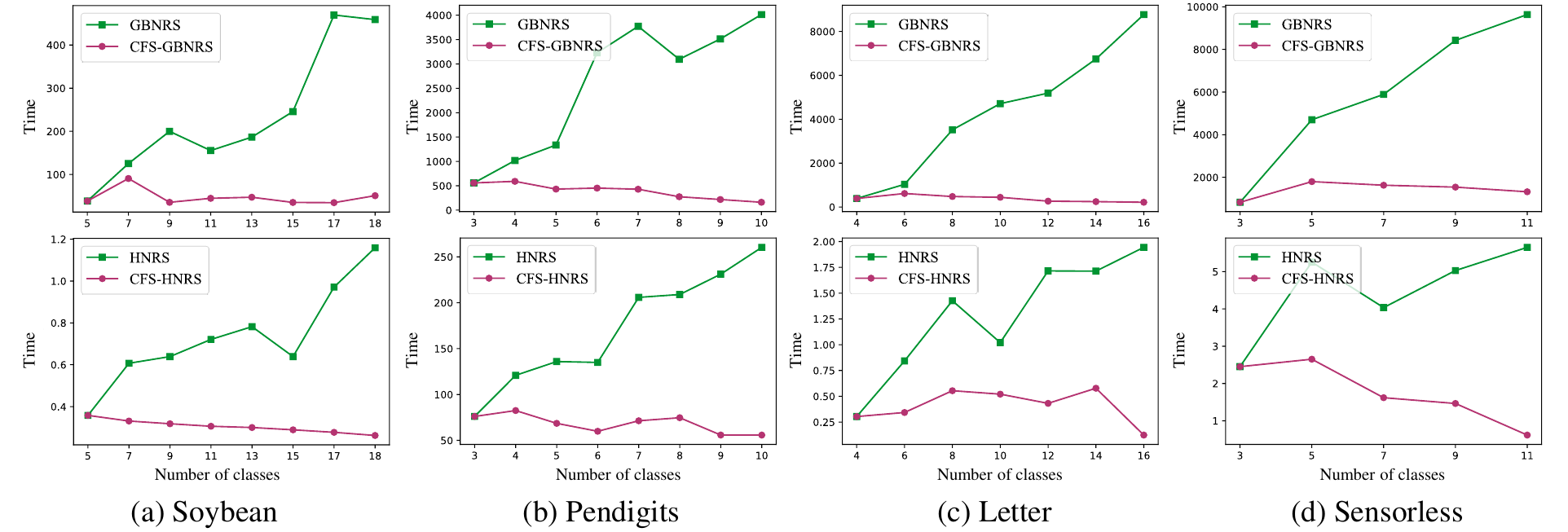}
\caption{Running time curves of the granular-ball method under our framework and the original granular-ball method as the number of classes increases.}
\label{fig: time}
\end{figure*}

The explosion of feature combinations poses a challenge for all rough set methods.
Although HNRS and GBNRS improve rough set efficiency through point set calculation, further optimization is needed in the open environment.
To address this, we introduced a knowledge base into our CFS-HNRS and CFS-GBNRS methods to enhance efficiency through knowledge transfer, avoiding redundant computations on existing data when new data arrives.
Therefore, in this subsection, we compared the average execution times of the original HNRS and GBNRS methods with our improved versions over multiple datasets.
We conducted ten repetitions of each experiment on four selected datasets: Soybean, Pendigits, Letter, and Sensorless, chosen for their ample instances and classes. 
Each dataset was initially processed with 30\% of its classes, gradually increasing by 10\% each period until all classes were covered.

The time comparison is shown in Fig. \ref{fig: time}.
The ordinate represents time in seconds, and the abscissa represents the number of classes in each period.
The figure intuitively shows that the running time of CFS-HNRS and CFS-GBNRS is significantly lower than that of HNRS and GBNRS, respectively, and the time consumption in different periods is very stable.
Because the original HNRS and GBNRS lack continuous feature selection capabilities, they must be run from scratch whenever new data is introduced.
Therefore, in comparison, the runtime time of the original HNRS and GBNRS methods increased significantly with the addition of new classes of data.
Specifically, our method saves about ten times the time cost of the original granular-ball method when each data set completes the run. %
In particular, for the GBNRS method, CFS-GBNRS can reduce the time overhead from nearly 10,000 seconds on the sensorless dataset to about 1,300 seconds.
Therefore, the CFS framework based on the knowledge base significantly improves the operating efficiency of the original granular-ball algorithms.

\section{Conclusion}
Feature selection is widely employed in data preprocessing. 
Current feature selection methods assume that all class labels are known in advance, neglecting the potential for unknown classes to appear in open environments.
This paper proposes a novel continual feature selection (CFS) framework that combines the advantages of continual learning and granular-ball computing, which consists of two learning stages.
The first stage builds a basic knowledge base through a multi-granularity representation of granular-ball to mitigate risks in open environments.
The second stage first uses the previously learned particle knowledge stage to identify unknowns, and then updates and consolidates the knowledge base for granular-ball knowledge transfer.
Furthermore, an optimal feature subset mechanism is designed to quickly select the optimal feature subset for the new period.
Finally, the effectiveness and efficiency of this framework are validated through experiments.
Future work includes more effectively constructing granular-balls that clearly delineate data distributions for identified unknown classes, and more efficiently integrating new and old knowledge within the knowledge base.



\ifCLASSOPTIONcaptionsoff
  \newpage
\fi

\bibliographystyle{IEEEtran}

\bibliography{reference}

\end{document}